\documentclass{article}

\usepackage[final]{corl_2023} 
\usepackage{amsthm, amsmath, amssymb}
\usepackage{graphics}
\usepackage{graphicx}
\usepackage{multirow}
\usepackage{caption}
\usepackage{subcaption}
\usepackage{enumitem}
\usepackage{hyperref}
\usepackage{wrapfig}

\usepackage{algorithm}
\usepackage{algpseudocode}
\usepackage{subcaption}
\usepackage{booktabs} 
\usepackage[tableposition=top]{caption}

\newtheorem{theorem}{Theorem}[section]
\newtheorem{lemma}[theorem]{\textbf{Lemma}}
\newtheorem{assumption}{Assumption}

\title{Marginalized Importance Sampling for Off-Environment Policy Evaluation}

\author{
  Pulkit Katdare\\
  Department of Electrical and Computer Engineering \\
  University of Illinois Urbana-Champaign
  Illinois, United States\\
  \texttt{katdare2@illinois.edu} \\
  \And 
  Nan Jiang \\
  Department of Computer Science \\
  University of Illinois Urbana-Champaign
  Illinois, United States
  \And
  Katherine Driggs-Campbell \\
  Department of Electrical and Computer Engineering \\
  University of Illinois Urbana-Champaign
  Illinois, United States
}

\begin{document}
\maketitle

\begin{abstract}
    Reinforcement Learning (RL) methods are typically sample-inefficient, making it challenging to train and deploy RL-policies in real world robots. Even a robust policy trained in simulation requires a real-world deployment to assess their performance. This paper proposes a new approach to evaluate the real-world performance of agent policies prior to deploying them in the real world. Our approach incorporates a simulator along with real-world offline data to evaluate the performance of any policy using the framework of Marginalized Importance Sampling (MIS). Existing MIS methods face two challenges: (1) large density ratios that deviate from a reasonable range and (2) indirect supervision, where the ratio needs to be inferred indirectly, thus exacerbating estimation error. Our approach addresses these challenges by introducing the target policy's occupancy in the simulator as an intermediate variable and learning the density ratio as the product of two terms that can be learned separately. The first term is learned with direct supervision and the second term has a small magnitude, thus making it computationally efficient. We analyze the sample complexity as well as error propagation of our two step-procedure. Furthermore, we empirically evaluate our approach on Sim2Sim environments such as Cartpole, Reacher, and Half-Cheetah. Our results show that our method generalizes well across a variety of Sim2Sim gap, target policies and offline data collection policies. We also demonstrate the performance of our algorithm on a Sim2Real task of validating the performance of a 7 DoF robotic arm using offline data along with the Gazebo simulator. 
\end{abstract}



\section{Introduction}
\vspace{-5pt}
Reinforcement Learning (RL) algorithms aim to select actions that maximize the cumulative returns over a finite time horizon. In recent years, RL has shown state-of-the-art performance over a range of complex tasks such as chatbots, \cite{chatgpt}, games \cite{ DBLP:journals/nature/SilverSSAHGHBLB17}, and robotics \cite{DBLP:journals/corr/abs-2203-01821, DBLP:journals/corr/abs-1910-07113, DBLP:conf/iros/YuKTL19, DBLP:conf/icra/KatdareLC22}. However, RL algorithms still require a large number of samples, which can limit their practical use in robotics~\cite{DBLP:conf/icml/HaarnojaZAL18, DBLP:journals/corr/SchulmanWDRK17}.

A typical approach is to train robust robot policies in simulation and then deploy them on the robot~\cite{DBLP:conf/iros/TobinFRSZA17, DBLP:conf/icra/DuWDAP21}. Such an approach, although useful, does not guarantee optimal performance on the real robot without significant fine tuning~\cite{DBLP:conf/icra/LeibovichJENT22}. In this work, we propose an approach that evaluates the real world performance of a policy using a robot simulator and offline data collected from the real-world~\cite{DBLP:conf/icra/KatdareLC22}. To achieve this, we employ the framework of off-policy evaluation (OPE) ~\cite{voloshin2019empirical}. 

OPE is the problem of using offline data collected from a possibly unknown \textit{behavior} policy to estimate the performance of a different \textit{target} policy. Classical OPE methods are based on the principle of importance sampling (IS)~\cite{precup2000eligibility, jiang2016doubly, levine2020offline}, which reweights each trajectory by its density ratio under the target versus the behavior policies. More recently, significant progress has been made on \textit{marginalized importance sampling} (MIS), which reweights transition tuples using the density ratio (or importance weights) over states instead of trajectories to overcome the so-called curse of horizon~\cite{liu2018breaking, nachum2020reinforcement, uehara2019minimax}. The density ratio is often learned via optimizing minimax loss functions. 

Most existing MIS methods are model-free, relying on data from the real environment to approximate the MIS weight function. However, a direct application of MIS methods to robotics carries two main disadvantages. (1) \textit{Large ratios:} MIS method learns distribution mismatch between the behavior and the target policies. When the mismatch between the behavior and the target policy is large, which is often the case, MIS method tend be challenging to learn. (2) \textit{Indirect supervision:} MIS methods requires no samples from target policies, which requires the weight being learned \textit{indirectly} via the Bellman flow equation. In states where coverage of the offline dataset is scarce, MIS methods tend to perform poorly. 

\begin{figure}[t]
    \centering
    \includegraphics[width=0.8\textwidth]{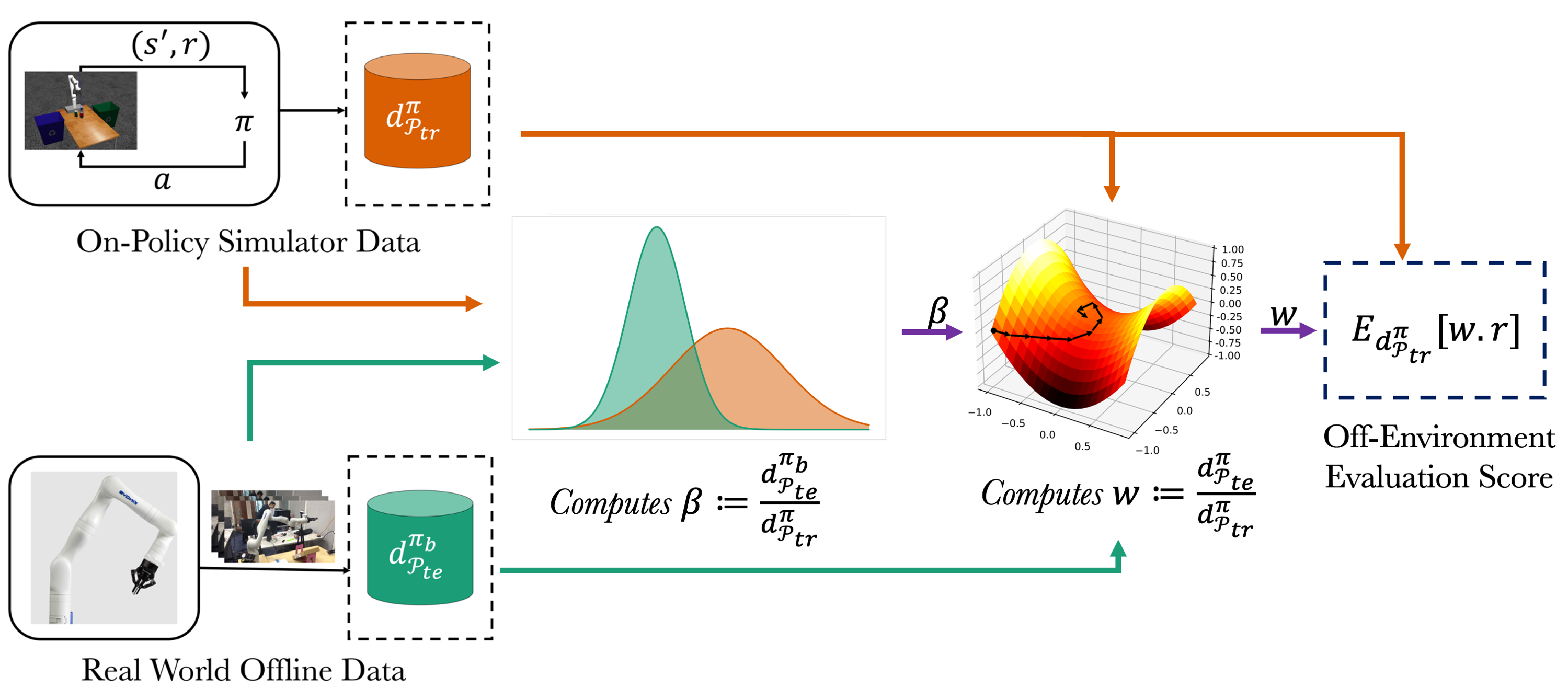}
    \caption{For a given policy $\pi$, we first collect on-policy data $d^{\pi}_{\mathcal{P}_{tr}}$ from a simulator environment. Using $d^{\pi}_{\mathcal{P}_{tr}}$ and offline data $d^\mathcal{D}_{\mathcal{P}_{te}}$ on the real world, we first calculate the importance sampling factor $\beta =d^{\pi_{b}}_{\mathcal{P}_{te}}/{d^{\pi}_{\mathcal{P}_{tr}}}$. This importance sampling factor essentially allows us to re-weight existing off-policy evaluation algorithm to estimate $w = {d^{\pi}_{\mathcal{P}_{te}}}/{d^{\pi}_{\mathcal{P}_{tr}}}$ which helps us estimate real-world performance of the agent $J_{\mathcal{P}_{te}}(\pi)$ using on-policy simulator data.}
    \label{fig:main_figure}
    \vspace{-10pt}
\end{figure}

In robotics, it is reasonable to assume access to a good but imperfect simulator of the real environment~\cite{brockman2016openai, coumans2021, 6386109}. In this work, we propose an improved MIS estimator that estimates the density ratio mismatch between the real world and the simulator. We further show that such a MIS estimator can be used to evaluate the real-world performance of a robot using just the simulator. As described in figure \ref{fig:main_figure}, we estimate the discrepancy between the real world and the simulator by using the target policy's occupancy in the simulator as an intermediate variable. This allows us to calculate the MIS weights as a combination of two factors, which can be learned independently. The first factor has direct supervision, while the second factor has a small magnitude (close to 1), thereby addressing both large ratios and indirect supervision issues mentioned above. We present a straightforward derivation of our method, examine its theoretical properties, and compare it to existing methods (including existing ways of incorporating a simulator in OPE) and baselines through empirical analysis. 

We make the following contributions. (1) We derive an MIS estimator for off-environment evaluation (Section~\ref{sec:weight estimator}). (2) We explore the theoretical properties of our off-environment evaluation estimator by proposing a sample-complexity analysis (Section~ \ref{sec:weight estimator}) and studying its special cases in linear and kernel settings. (3) We empirically evaluate our estimator on both Sim2Sim as well as Sim2Real environments (Section~\ref{sec:experiment}). For the Sim2Sim experiments, we perform a thorough ablation study over different Sim2Sim gap, data-collection policies and target policies, environments (Taxi, Cartpole, Reacher, and Half-Cheetah). Furthermore, we demonstrate practicality of our approach on a sim2real task by validating performance of a Kinova robotic arm over using offline data along with Gazebo based Kinova simulator. 

\section{Preliminaries} \label{sec:prelim}
Robot learning problems are often modelled as an infinite-horizon discounted Markov Decision Process (MDP). MDP is specified by $(\mathcal{S}, \mathcal{A}, P, R, \gamma, d_0)$. Here, $\mathcal{S}$ and $\mathcal{A}$ are the state and the action spaces, $P: \mathcal{S}\times\mathcal{A}\to \Delta(\mathcal{S})$ is the transition function ($\Delta(\cdot)$ is the probability simplex). We also define the reward function $R:\mathcal{S}\times\mathcal{A}\to \Delta([0, R_{max}])$,  $\gamma \in [0, 1)$ is the discount factor, and $d_0 \in \Delta(\mathcal{S})$ is the initial state distribution. A policy $\pi: \mathcal{S}\to\Delta(\mathcal{A})$ 
induces a distribution of trajectory: $\forall t \ge 0$,
$s_0 \sim d_0, a_t \sim \pi(\cdot | s_t), r_t \sim R(\cdot | s, a), s_{t+1} \sim P(\cdot| s_t, a_t).$  
The performance of $\pi$ is measured by its expected discounted return under the initial state distribution, defined as
$J_P(\pi) = (1-\gamma) \mathbb{E}[\sum_{t=0}^\infty \gamma^t r_t | \pi, d_0]$; here, we use the subscript $P$ in $J_P(\pi)$ to emphasize the dynamics w.r.t.~which the return is defined, since we will consider both the true environment and the simulator in the rest of this paper and the subscript will help distinguish between them.  $J_P(\pi)$ also has an alternative expression  $J_{P}(\pi) := \mathbb{E}_{(s, a) \sim d^\pi_P, r \sim R(s, a)}[r]$, where 
\begin{align}\label{eq:rlform}
\textstyle    d^\pi_P(s, a) =  (1-\gamma)\sum_{t=0}^\infty \gamma^t \mathbb{P}[s_t = s, a_t = a | \pi, d_0] 
\end{align}
is the discounted state-action occupancy induced by $\pi$ from $d_0$.  An important quantity associated with a policy is its Q-function $Q^\pi_{P}$, which 
satisfies the Bellman equation $Q_P^\pi(s,a) = \mathbb{E}_{r \sim R(s,a), s' \sim P(s,a)}[r + \gamma Q_P^\pi(s', \pi)]$, where $f(s',\pi):=\mathbb{E}_{a'\sim \pi(\cdot|s')}[f(s',a')]$. We can also define the state-value function $V_P^\pi(s) = Q_P^\pi(s, \pi)$, and $J(\pi) = (1-\gamma) \mathbb{E}_{s \sim d_0}[V_P^\pi(s)]$. \\
\textbf{OPE and Marginalized Importance Sampling:} In off-policy evaluation (OPE), we want to evaluate a target policy $\pi$ using data collected from a different policy in the real environment, denoted by its dynamics $P$. As a standard simplification, we assume data is generated i.i.d.~as $(s,a) \sim \mu, r \sim R(s,a), s'\sim P(s,a)$, and the sample size is $n$. When the data is generated from some behavior policy $\pi_b$, $\mu$ can correspond to its occupancy $d_{P}^{\pi_b}$. We will use $\mathbb{E}_{\mu}[\cdot]$ as a shorthand for taking expectation over $(s,a,r,s')$ generated from such a data distribution in the real environment. 

The key idea in marginalized importance sampling (MIS) is to learn the weight function $w_P^{\pi/\mu}(s,a):=\frac{d_P^\pi(s,a)}{\mu(s,a)}$. 
Once this function is known, $J(\pi)$ can be estimated as 
$J(\pi) =\mathbb{E}_{(s, a) \sim d^{\pi}_{P}, r \sim R(s, a)}[r] = \mathbb{E}_\mu[w_P^{\pi/\mu}(s,a) \cdot r].$ 
Note that $\mathbb{E}_{\mu}[\cdot]$ can be empirically approximated by the dataset sampled from the real environment. The real challenge in MIS is how to learn $w_{P}^{\pi/\mu}$. Existing works often do so by optimizing minimax loss functions using Q-functions as discriminators, and is subject to both difficulties (large ratios and indirect supervision) mentioned in the introduction. We refer the readers to \cite{jiang2020minimax} for a summary of typical MIS methods. \\
\textbf{Learning Density Ratios from Direct Supervision:} Given two distributions $p$ and $q$ over the same space $\mathcal{X}$, the density ratio $p(x)/q(x)$ can be learned \textit{directly} if we have access to samples from both $p$ and $q$, using the method proposed by \cite{DBLP:journals/corr/abs-0809-0853}: 
\begin{align}\label{eq:beta_optimizer}
    \frac{p(x)}{q(x)} = \arg \max_{f: \mathcal{X}\to\mathbb{R}_{>0}} \mathbb{E}_{x \sim p}[\ln f(x)] - \mathbb{E}_{x \sim q}[f(x)] + 1.
\end{align}
To guarantee generalization over a finite sample when we approximate the expectations empirically, we will need to restrict the space of $f$ that we search over to function classes of limited capacities, such as RKHS or neural nets, and $p(x)/q(x)$ can still be well approximated as long as it can be represented in the chosen function class (i.e., \textit{realizable}). More concretely, if we have $n$ samples $x_1, \ldots, x_n$ from $p$ and $m$ samples $\tilde{x}_1, \ldots, \tilde{x}_m$ from $q$, and use $\mathcal{F}$ to approximate $p(x)/q(x)$, the learned density ratio can be made generalizable by adding a regularization term $I(f)$ to improve the statistical and computational stability of learning:
\begin{align} \label{eq:nguyen}
\arg \max_{f \in \mathcal{F}} \frac{1}{n}\sum_{i} \ln f(x_i) - \frac{1}{m}\sum_{j}f(\tilde{x}_j) + \frac{\lambda}{2} I(f)^2.
\end{align}
\vspace{-5pt}
\section{Related Work} \label{sec:related}
\subsection{Reinforcement Learning applications in Robotics}
There are three different themes that arise in reinforcement learning for robotics~\cite{sikchi2023imitation, DBLP:journals/corr/abs-2001-01866}. 
(1) \textbf{Sim2Real:} algorithms are primarily concerned with learning robust policy in simulation by training the algorithms over a variety of simulation configurations ~\cite{DBLP:journals/tase/HoferBHGMGAFGLL21, DBLP:conf/iros/TobinFRSZA17, DBLP:journals/corr/abs-1903-11774}. Sim2Real algorithms, although successful, still require a thorough real-world deployment in order to gauge the policy’s performance. 
(2) \textbf{Imitation learning} algorithms learn an optimal policy by trying to mimic offline expert demonstrations ~\cite{DBLP:journals/jmlr/RossGB11, DBLP:conf/nips/HoE16, DBLP:journals/corr/abs-1901-09387}. Many successful imitation learning algorithms minimize some form of density matching between the expert demonstrations and on-policy data to learn optimal policy. A key problem with imitation learning is the fact that it requires constant interaction with the real-world environment in order to learn an optimal policy. (3) \textbf{Offline reinforcement learning} is a relatively new area. Here the idea is to learn an optimal policy using offline data without any interaction with the environment ~\cite{kumar2020conservative, DBLP:journals/corr/abs-2106-10783, DBLP:conf/nips/YuKRRLF21, DBLP:conf/aaai/YuYSNE23}. Offline reinforcement learning has recently demonstrated performance at-par with classical reinforcement learning in a few tasks. However, offline reinforcement learning algorithms tend to overfit on the offline data. Thus, even offline learning modules too require an actual deployment in-order to assess performance. 
\subsection{Off-Policy Evaluation}
\vspace{-5pt}
We review related works in this section, focusing on comparing to existing OPE methods that can leverage the side information provided by an imperfect simulator.
(1) \textbf{Marginalized Importance Sampling (MIS):} MIS methods tend to assume the framework of data collection and policy evaluation being on the same environment. To that end, there are both model-free ~\cite{DBLP:journals/corr/abs-1906-04733, zhang2020gradientdice, zhang2019gendice} and model-based variants of MIS methods and face the aforementioned two challenges (large magnitude of weights and indirection supervision) simultaneously. Model based variants of MIS sometimes tend to be doubly robust (DR) in nature ~\cite{uehara2019minimax, tang2019doubly} and can benefit from Q-functions as control variates, which can be supplied by the simulator. However, the DR version of MIS is a meta estimator, and the weight $d_{P_{te}}^\pi/\mu$ still needs to be estimated via a ``base'' MIS procedure. Therefore, the incorporation of the simulator information does not directly address the challenges we are concerned with, and there is also opportunity to further combine our estimator into the DR form of MIS. 
(2) \textbf{Model-based methods:} Model-based estimators first approximate the transition dynamics of the target environment~\cite{fu2021benchmarks}, which is further used to generate rollouts and evaluate performance for any target policy. One way of incorporating a given imperfect simulator in this approach is to use the simulator as ``base predictions,'' and only learn an additive correction term, often known as residual dynamics~\cite{voloshin2021minimax}. This approach combines the two sources of information (simulator and data from target environment) in a very different way compared to ours, and are more vulnerable to misspecification errors than model-free methods. 

\vspace{-5pt}
\section{Weight Estimator}\label{sec:weight estimator}
\vspace{-5pt}
Recall that our goal is to incorporate a given simulator into MIS. We will assume that the simulator shares the same $\mathcal{S}, \mathcal{A}, \gamma, d_0$ with the real environment, but has its own transition function $P_{tr}$ which can be different from $P_{te}$. As we will see, extension to the case where the reward function is unknown and must be inferred from sample rewards in the data is straightforward, and for simplicity we will only consider difference in dynamics for most of the paper. \\
\textbf{Split the weight:} The key idea in our approach is to split the weight $w_P^{\pi/\mu}$ into two parts 
by introducing $d_{P_{tr}}^\pi$ as an intermediate variable:
\begin{align*}
&w_{P}^{\pi/\mu}(s,a) = \frac{d_{P_{te}}^\pi(s,a)}{\mu(s,a)} =\underbrace{\frac{d_{P_{tr}}^\pi(s,a)}{\mu(s,a)}}_{\substack{:= \beta \\ \textrm{(direct supervision)}}} \cdot \underbrace{\frac{d_{P_{te}}^\pi(s,a)}{d_{P_{tr}}^\pi(s,a)}}_{\substack{:= w^{\pi}_{P_{te}/P_{tr}}\textrm{(magnitude $\simeq 1$)}}}
\end{align*}
Note that $d_{P_{tr}}^\pi$ is the occupancy of $\pi$ in the simulator, which we have free access to. The advantage of our approach is that by estimating $\beta$ and $w_{P_{te}/P_{tr}}^\pi$ separately, we avoid the situation of running into  the two challenges mentioned before simultaneously, and instead address one at each time: $\beta = d_{P_{tr}}^{\pi} / \mu$ has large magnitude but can be learned directly via \cite{DBLP:journals/corr/abs-0809-0853} without the difficult minimax optimization typically required by MIS, and we expect $w^\pi_{P_{te}/P_{tr}} = d_{P_{te}}^\pi/d_{P_{tr}}^\pi$ to be close to $1$ when $P_{te} \approx P_{tr}$ (and thus easier to learn). \\
\textbf{Estimate $w^\pi_{{P_{te}/P_{tr}}}$:} Since $\beta$ is handled by the method of \cite{DBLP:journals/corr/abs-0809-0853}, the key remaining challenge is how to estimate $w^\pi_{{P_{te}/P_{tr}}}$. (Interestingly, $\beta$ also plays a key role in estimating $w^\pi_{{P_{te}/P_{tr}}}$, as will be shown below.) Note that once we have approximated $w^\pi_{{P_{te}/P_{tr}}}$ with some $w$, we can directly reweight the state-action pairs from the simulator (i.e., $d_{P_{tr}}^\pi$) if the reward function is known (this is only assumed for the purpose of derivation), i.e., 
\begin{align*}
J_{P_{te}}(\pi) \approx \mathbb{E}_{(s,a) \sim d_{P_{tr}}^\pi, r \sim R(s,a)}[w \cdot r],
\end{align*}
and this becomes an identity if $w = w^\pi_{{P_{te}/P_{tr}}}$. Following the derivation in \cite{uehara2019minimax, jiang2020minimax}, we now reason about the error of the above estimator for an arbitrary $w$ to derive an upper bound as our loss for learning $w$: 
\begin{align}\label{eq:derivation}
&~ |\mathbb{E}_{(s,a) \sim d_{P_{tr}}^\pi, r \sim R(s,a)}[w \cdot r] - J_{P_{te}}(\pi)|  \nonumber\\
= &~ |\mathbb{E}_{(s,a) \sim d_{P_{tr}}^\pi, s' \sim P(s,a)}[w \cdot (Q_{P_{te}}^\pi(s,a) -\gamma Q_{P_{te}}^\pi(s', \pi))] - (1-\gamma) \mathbb{E}_{s \sim d_0}[Q_{P_{te}}^\pi(s, \pi)]|  \nonumber\\
\le &~ \sup_{q \in \mathcal{Q}} |\mathbb{E}_{d_{P_{tr}}^\pi \times P_{te}}[w \cdot (q(s,a) -\gamma q(s', \pi))] - (1-\gamma) \mathbb{E}_{s \sim d_0}[q(s, \pi)]|. 
\end{align}
Here $d_{P_{tr}}^\pi \times P_{te}$ is a shorthand for $(s,a) \sim d_{P_{tr}}^\pi, s'\sim P(s,a)$. 
In the last step, we handle the unknown $Q_{P}^\pi$ by a relaxation similar to \cite{liu2018breaking, uehara2019minimax, jiang2020minimax}, which takes an upper bound of the error over $q\in\mathcal{Q}$ for some function class $\mathcal{Q} \subset \mathbb{R}^{\mathcal{S}\times\mathcal{A}}$, and the inequality holds as long as $Q_P^\pi \in conv(\mathcal{Q})$ with $conv(\cdot)$ being the convex hull.\\ 
\textbf{Approximate $d_{P_{tr}}^\pi \times P_{te}$:}
The remaining difficulty is that we will need samples from $d_{P_{tr}}^\pi \times P_{te}$, i.e., $(s,a)$ sampled from $\pi$'s occupancy in the $P_{tr}$ \textit{simulator}, and the next $s'$ generated in the $P_{te}$ \textit{real environment}. While there is no natural dataset for such a distribution,  we can take the data from the real environment, $(s,a,s') \sim \mu \times P_{te}$, and reweight it using $\beta = d_{P_{tr}}^\pi / \mu$ to approximate expectation w.r.t.~$d_{P_{tr}}^\pi \times P_{te}$, i.e.,
\begin{align*}
& (s,a,s') \sim d_{P_{tr}}^\pi \times P_{te} \quad \Longleftrightarrow (s,a,s') \sim \mu \times P_{te}  ~ \text{reweighted with} ~ \beta:= d_{P_{tr}}^\pi/ \mu.
\end{align*}
Based on such an observation, we can further upper-bound $|\mathbb{E}_{(s,a) \sim d_{P_{tr}}^\pi, r \sim R(s,a)}[w \cdot r] - J_{P_{te}}(\pi)|$ from end of Equation~\ref{eq:derivation} with:
\begin{align} \label{eq:simple_loss}
& \sup_{q \in \mathcal{Q}} L_w(w, \beta, q):=  |\mathbb{E}_{\mu}[w \cdot \beta \cdot (q(s,a) -\gamma q(s', \pi))] - (1-\gamma) \mathbb{E}_{s \sim d_0}[q(s, \pi)]| .
\end{align}
As our derivation has shown, this is a valid upper bound of the error as long as $ conv(\mathcal{Q})$ can represent $Q_{P_{te}}^\pi$. We also need to show that the upper bound is non-trivial, i.e., when $w = w^\pi_{{P_{te}/P_{tr}}}$, the upper bound should be $0$. 
This is actually easy to see, as for any $q$:
\begin{align}\label{eq:uniqueness}
&~ L(w^\pi_{P_{te}/P_{tr}}, \beta, q):= 
|\mathbb{E}_{d_{P_{tr}}^\pi \times P_{te}}[w^\pi_{P_{te}/P_{tr}}\cdot (q(s,a) -\gamma q(s', \pi))]  - (1-\gamma) \mathbb{E}_{s \sim d_0}[q(s, \pi)]| \\
&= |\mathbb{E}_{d_{P_{te}}^\pi \times P}[q(s,a) -\gamma q(s', \pi)] - (1-\gamma) \mathbb{E}_{s \sim d_0}[q(s, \pi)]| = 0.
\end{align}
The last step directly follows from the fact that $d_{P_{te}}^\pi$ is a valid discounted occupancy and obeys the Bellman flow equation. Therefore, it makes sense to search for $w$ over a function class $\mathcal{W} \subset \mathbb{R}^{\mathcal{S}\times\mathcal{A}}$ to minimize the loss $\sup_{q\in\mathcal{Q}} L(w, \beta, q)$. \\
\textbf{Final estimator:} To summarize our estimation procedure, we will first use \cite{DBLP:journals/corr/abs-0809-0853} to estimate $\hat \beta \approx d_{P'}^\pi /\mu$ with a function class $\mathcal{F}$, and plug the solution into our loss for estimating $w^\pi_{P_{te}/P_{tr}}$, i.e.,
\begin{align}\label{eq:optimizer}
\hat w = \arg \min_{w \in \mathcal{W}} \sup_{q\in\mathcal{Q}} L_w(w, \hat \beta, q). 
\end{align}
As mentioned above, if the reward function is known, we can use $\mathbb{E}_{(s,a) \sim d_{P_{tr}}^\pi, r \sim R(s,a)}[\hat w \cdot r]$ as our estimation of $J_{P_{te}}(\pi)$. We can also demonstrate interesting properties of our optimization like the effect of a linear function class and RKHS function class. This discussion is deferred to the supplementary materials section \ref{sec:case_studies}.

\textbf{Sample Complexity Guarantee:} We can further provide an upper-bound on the performance of our final estimator under the following two assumptions. 
\vspace{-5pt}
\begin{assumption}[Boundedness]\label{ass: bounds}
We assume $\forall f \in \mathcal{F}$, $0 < {C}_{\mathcal{F}, \min} \le f \le {C}_{\mathcal{F}, \max}$. Define $C_\mathcal{F} := C_{\mathcal{F}, \max} + \max(\log C_{\mathcal{F}, \max}, -\log C_{\mathcal{F}, \min})$. Similarly, $\forall w\in\mathcal{W}$, $w\in [0, C_{\mathcal{W}}]$, and $\forall q\in\mathcal{Q}$, $q\in[0, C_{\mathcal{Q}}]$. 
\end{assumption}
\begin{assumption}[Realizability of $\mathcal{F}$]\label{ass: existence}
$d^\pi_{P'}/\mu \in \mathcal{F}$.
\end{assumption}
\begin{theorem}\label{thm:finite-sample}
Let $\hat \beta$ be our approximation of $d^\pi_{P'}/\mu$ which we found using \cite{DBLP:journals/corr/abs-0809-0853}. We utilize this $\hat{\beta}$ to further optimize for $\hat{w}_n$ (equation \ref{eq:optimizer}) using n samples. In both cases, $\mathbb{E}_{(s,a) \sim d_{P_{tr}}^\pi}[\cdot]$ is also approximated with $n$ samples from the simulator $P_{tr}$. 
Then, under Assumptions~\ref{ass: bounds} and \ref{ass: existence} along with the additional assumption that $Q^\pi_{P_{te}} \in C(\mathcal{Q})$ with probability at least $1-\delta$, the Off Environment Evaluation error can be bounded as 
\begin{align}
 \begin{split}
     &|\mathbb{E}_{(s, a) \sim d^\pi_{P_{tr}}, r \sim R(s, a)} [\hat{w}_n \cdot r ] - J_{P_{te}}(\pi)|  \leq \min_{w \in \mathcal{W}} \max_{q \in \mathcal{Q}} |L_w(w, \beta, q)| \\
     &+ 2 C_{\mathcal{W}} \cdot C_\mathcal{Q}\cdot \tilde{O}\left(\sqrt{ \left\|\frac{d^\pi_{P'}}{\mu} \right\|_\infty \cdot \left(4\mathbb{E} \mathcal{R}_n(\mathcal{F}) + C_{\mathcal{F}}\sqrt{\frac{2 \log(\frac{2}{\delta})}{n}}\right)}\right) \\
     &+ 2 \mathcal{R}_n(\mathcal{W}, \mathcal{Q}) + C_{\mathcal{Q}} C_{\mathcal{W}}\sqrt{\frac{\log (\frac{2}{\delta})}{2n}} 
     \end{split}
 \end{align}
where $\mathcal{R}_n(\mathcal{F}), \mathcal{R}_n(\mathcal{W}, \mathcal{Q})$ are the Radamacher complexities of function classes $\{(x, y) \rightarrow f(x) - \log(f(y)) : f \in\mathcal{F}\}$  and 
$\{(s, a, s') \rightarrow (w(s, a) \cdot \frac{d^\pi_{P'}(s, a)}{\mu(s, a)} \cdot (q(s, a) - \gamma q(s', \pi)) : w\in\mathcal{W}, q\in\mathcal{Q}\} $, respectively, 
$\|d^\pi_{P'}/\mu\|_\infty := \max_{s,a} d^\pi_{P'}(s,a) / \mu(s,a)$ measures the distribution shift between $d^\pi_{P'}$ and $\mu$, 
and $\tilde O(\cdot)$ is the big-Oh notation suppressing logarithmic factors. 
\end{theorem}
Note that we do not make realizability assumption for $\mathcal{W}$ in the theorem above. Realizability assumption is reflected in the $\inf_{w\in \mathcal{W}} \sup_{q\in\mathcal{Q}} |L(w, \beta, q)|$, which equals $0$ when $d^\pi_{P_{te}}/d^\pi_{P_{tr}} \in \mathcal{W}$. The remaining terms vanishes to $0$ at an $O(1/\sqrt{n})$ rate when $n \to \infty$. \\
\textbf{Generalizing Off-Environment Policy Evaluation:} 
A key advantage of our two-step approach is that we can improve many existing off-policy evaluation algorithm with a similar two-step process. In this work, we use our two-step procedure with GradientDICE---which is an empirically state-of-the-art estimator in the DICE family---can also be similarly adapted as below, which we use in our experiments. Detailed derivation for the same can be found in the supplementary materials \ref{sec:beta_gradient_dice}.

\begin{table}[b]
\captionsetup{singlelinecheck = false, justification=justified}
\captionof{table}{Log mean squared error between the performance predicted by our $\beta$-DICE algorithm and the real world performance of the robot. We observe that our method is able to outperform DICE based baselines by a comfortable margin.}
\normalsize
\label{tab:kinova}
\begin{center} 
\begin{tabular}{lccccc}
\toprule 
 & \multicolumn{5}{c}{\textbf{Log\textsubscript{10} Mean Squared Error} ($\downarrow$)} \\
Algorithm&Kinova (Sim2Real) & Taxi & Cartpole & Reacher & Half-Cheetah\\
\midrule
Simulator & -3.96 & -0.19 & -2.58 &-1.09  & 1.18 \\
$\beta$-DICE (Ours) & \textbf{-4.38}&  \textbf{-1.60} & \textbf{-4.19} &\textbf{-4.08} & \textbf{-3.42} \\
GenDICE & -3.48 & -0.13 & -2.84 & -2.61 & -2.96 \\
GradientDICE & -3.49 & -0.59 & -1.45 & -3.17 & -2.16 \\
DualDICE & -3.48 & -0.48 & -0.99 &-2.88 & -2.12 \\
\bottomrule
\end{tabular}
\end{center}
\vspace{-15pt}
\end{table}
\begin{figure}
\centering
\begin{subfigure}{\textwidth}
    \centering
    \includegraphics[scale=0.5]{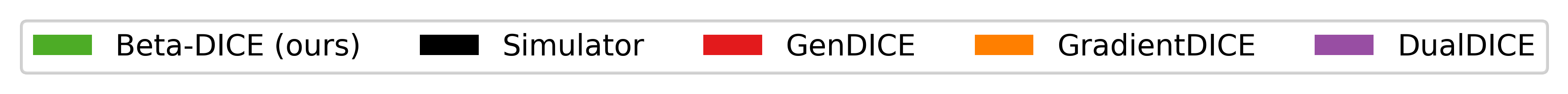}
    \label{fig:cartpole_legend}
\end{subfigure}\hfill
\begin{subfigure}{\textwidth}
    \centering
    \includegraphics[width=0.82\textwidth]{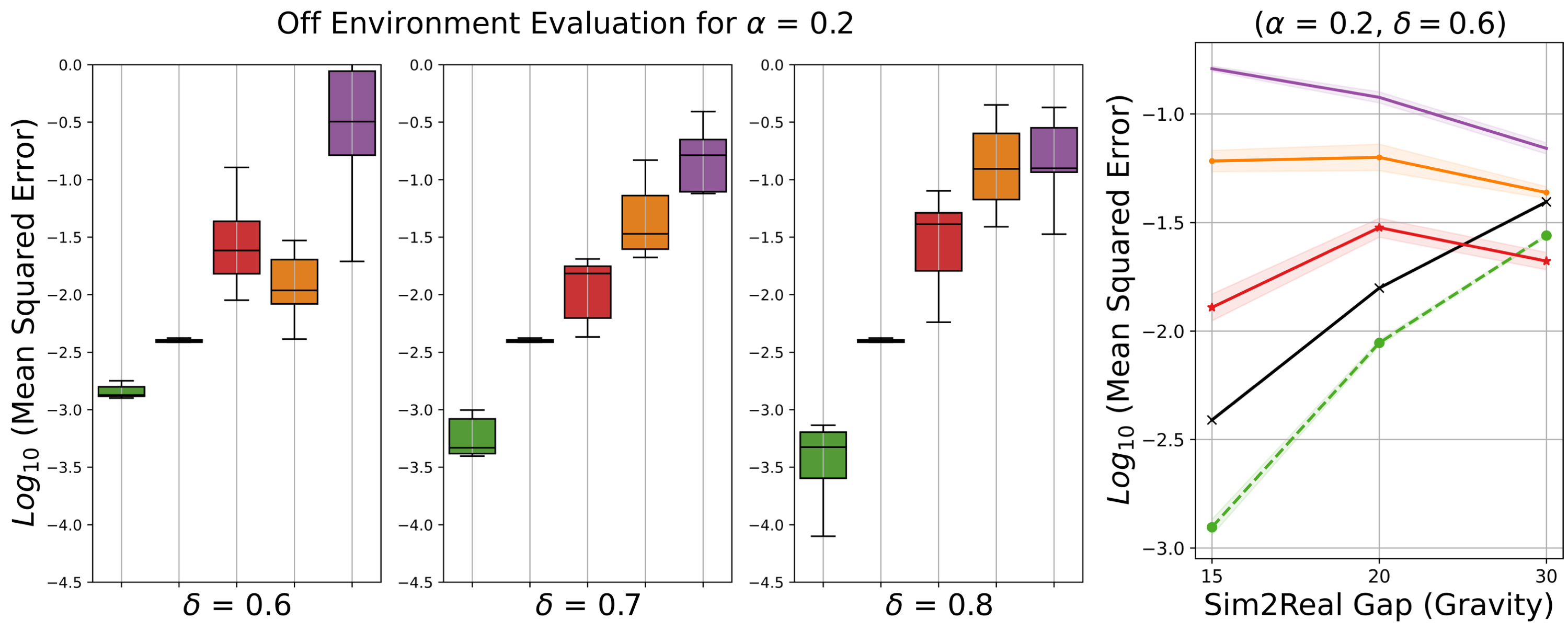}
    \caption{Target Policy $\alpha = 0.2$}
    \label{fig:cartpole}
\end{subfigure}
\begin{subfigure}{\textwidth}
    \centering
    \includegraphics[width=0.82\textwidth]{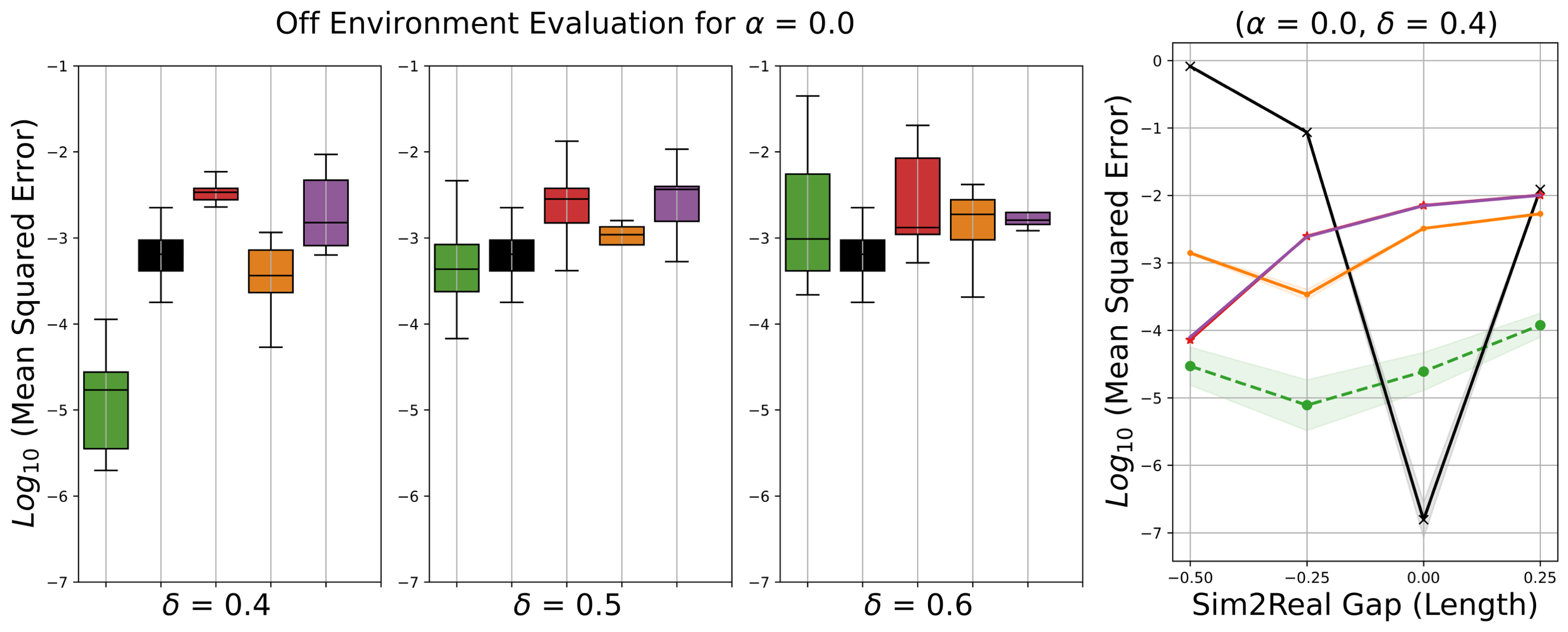}
    \caption{Target Policy $\alpha = 0.0$}
    \label{fig:reacher}
\end{subfigure}\hfill
    \caption{We demonstrate the effectiveness of $\beta$-DICE on Cartpole (a) and Reacher (b) Sim2Sim environment. For Cartpole environment we demonstrate the performance of $\beta$-DICE over for a Sim2Sim pair of $\{10, 15\} m/s^2$. Similarly for Reacher the Sim2Sim pair is (0.1m, 0.075m) for the length of the link.  On left hand side, we demonstrate the effect of $\beta$-DICE with different data collection policies while keeping the target policies fixed. On the right hand side, we demonstrate the impact of $\beta$-DICE with increasing Sim2Sim gap keeping the offline data collection policy the same. We observe that our $\beta$-DICE algorithm comfortably outperforms closest DICE based baselines.} 
\end{figure}

\section{Experiments} \label{sec:experiment} 
\subsection{Sim2Sim Validation of $\beta$-DICE}
\begin{figure}
\begin{subfigure}{\textwidth}
    \centering
    \includegraphics[scale=0.6]{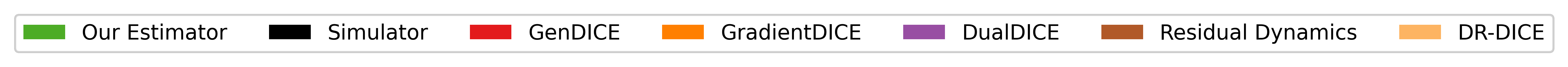}
\end{subfigure}\hfill
\begin{subfigure}{\textwidth}
    \centering
    \includegraphics[width=0.8\textwidth]{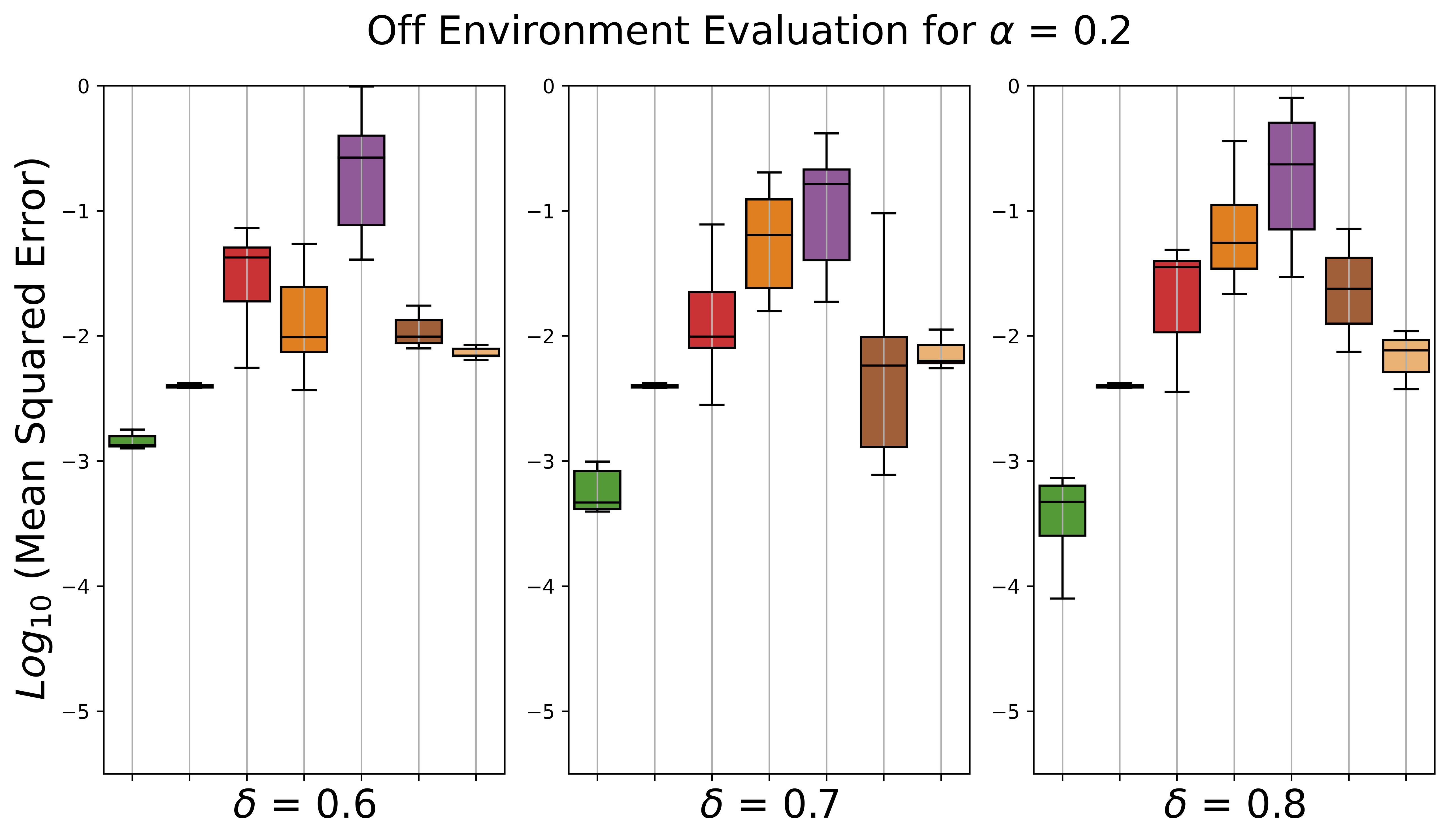}
    \caption{$\alpha=0.2$}
\end{subfigure}\hfill
\begin{subfigure}{\textwidth}
\centering
    \includegraphics[width=0.8\textwidth]{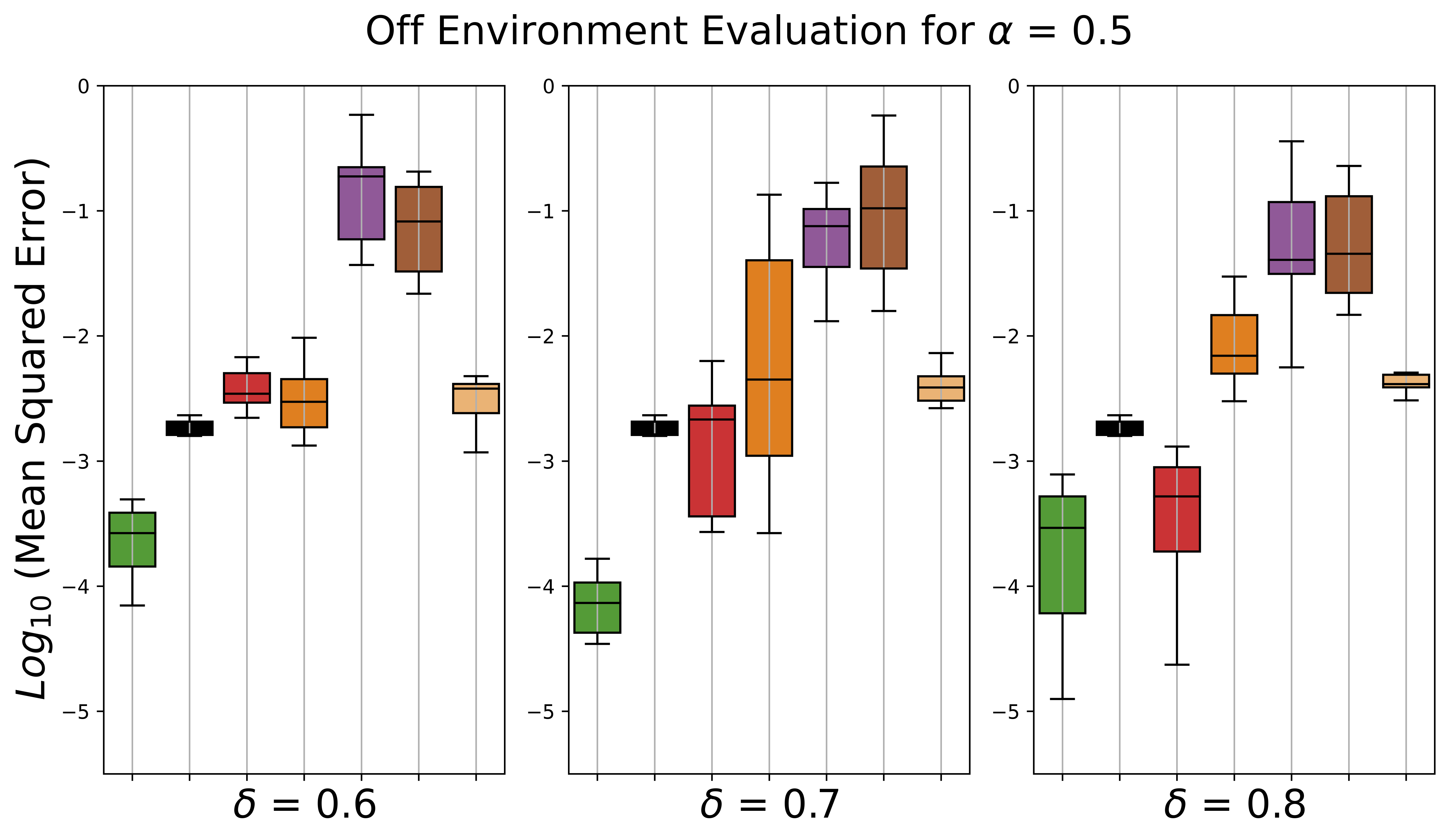}
    \caption{$\alpha=0.5$}
\end{subfigure}
\begin{subfigure}{\textwidth}
\centering
    \includegraphics[width=0.8\textwidth]{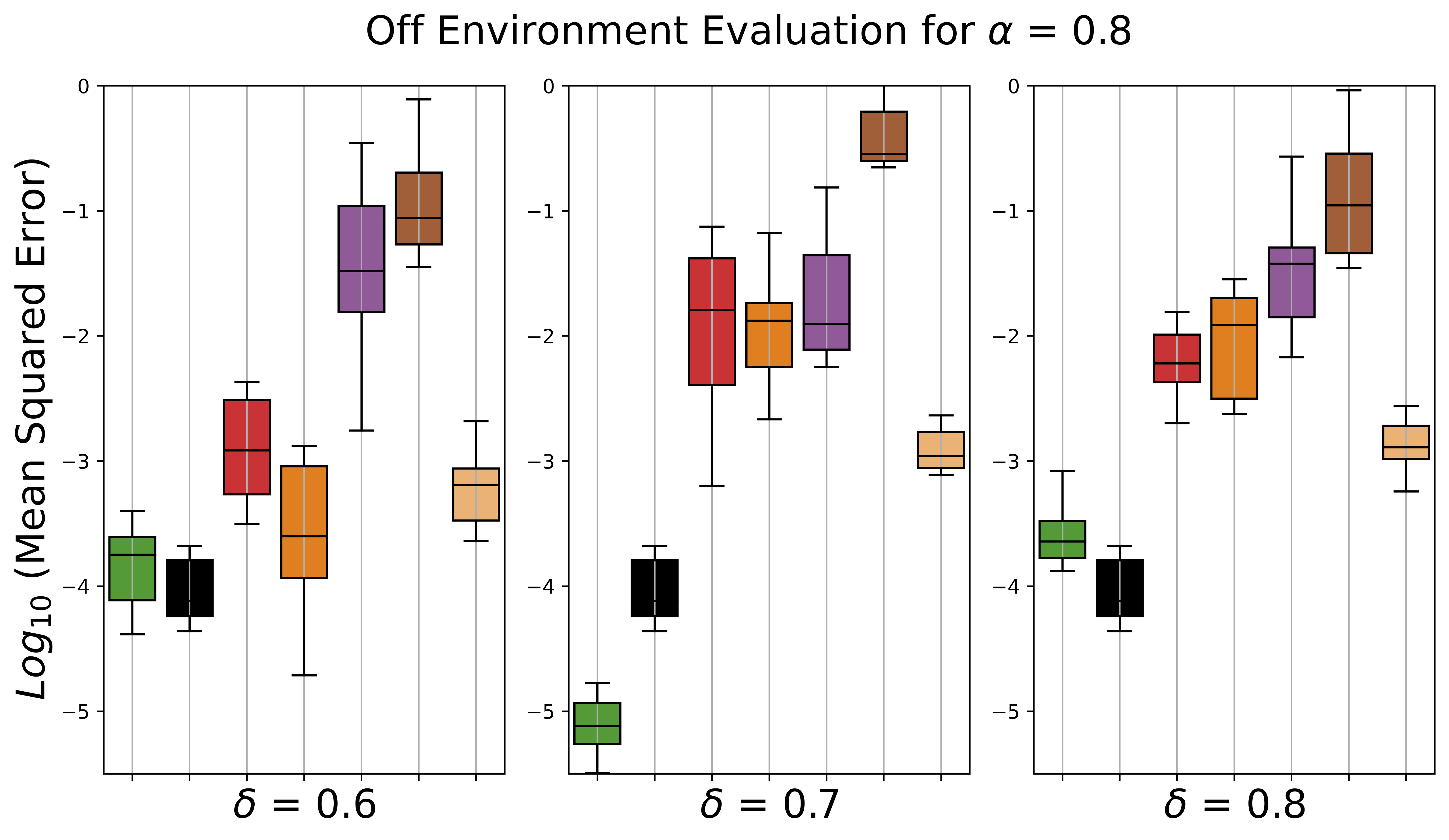}
    \caption{$\alpha=0.8$}
\end{subfigure}
    \caption{Evaluation error for the Cartpole environment over varying behavior policies $\delta =\{0.6, 0.7, 0.8\}$ and a fixed target policy using $\beta$-DICE for the cartpole environment with additional baselines. For these set of experiments the training environment has gravity = $10 m/s^2$, while the test environment has gravity = $15.0 m/s^2$. In (a), (b), (c) the target policies that we use are $\alpha = \{0.2, 0.5, 0.8$\}}
    \label{fig:cartpole_extra_baselines}
\end{figure}

\vspace{-5pt}
\textbf{Experimental Setting:} In the Sim2Sim experiments, we aim to show the effectiveness of our approach across different target policies, offline dataset as well as changing sim2sim gap. We further show the effectiveness of our approach over different types of Sim2Sim environments like Taxi (Tabular), Cartpole (discrete-control), Reacher (continuous control), and HalfCheetah (continuous control) environments. For each of these environments, we refer to the default configurations of these environments as the simulator environment. We further create a “real” world environment by changing key configurations from each of these environments. For example, we modify the transition probability in taxi, gravity in cartpole, and link lengths in reacher. These kinds of configuration changes help us assess the performance limits of our algorithm across a variety of sim2sim gap. In table \ref{tab:sim2sim_parameters}, we list all the different sim2sim environments configurations over which we experimented our algorithm. Typically these configurations are such that the real-world performance predicted by the simulator alone is off by 9-45\%  \\
We first collect our offline data by using a noisy pre-trained policy which is parameterised by $\delta$. Higher the $\delta$, noisier the data-collection policy. For Taxi and Cartpole environment, we using a uniform random policy for the noise, while we choose zero-mean gaussian policy for continuous environments like reacher and halfcheetah. Using this offline data as well as our simulator, we now evaluate the performance of any target policy, which we parameterise by $\alpha$. Target policy is further defined by a mixture of another pre-trained policy with noise. More the $\alpha$, more the randomness in the policy. Detailed experimental details along with the setup has been detailed in Appendix \ref{sec:experimental_details}
\\
\textbf{Results and Observations:} We present the detailed results for all the four environments in figures \ref{fig:additional_taxi} (Taxi), figures \ref{fig:cartpole}, \ref{fig:additional_cartpole} (Cartpole), figures \ref{fig:reacher} and \ref{fig:additional_reacher} (Reacher) and figure \ref{fig:halfcheetah} (HalfCheetah). For the boxplot, we fix target policy ($\alpha$) and demonstrate the evaluation error for our algorithm across a range of offline dataset ($\delta$) while keeping Sim2Sim gap fixed. For the line plots, we demonstrate the effectiveness of our algorithm across a changing Sim2Sim gap, while keeping the offline data ($\delta$) and target policy $\alpha$ fixed. We also compare our algorithm against DICE baselines GradientDICE~\cite{zhang2020gradientdice}, GenDICE~\cite{zhang2019gendice}, DualDICE~\cite{DBLP:journals/corr/abs-1906-04733}. DICE baselines are currently the state-of-the-art algorithm in off-policy evaluation and are known to outperform even hybrid off-policy evaluation algorithms. In figure \ref{fig:cartpole_extra_baselines}, we also compare our algorithm against hybrid off-policy evaluation baselines for the cartpole environment (further details in section~\ref{sec:experimental_details}). We observe that our method is able to comfortably outperform closest DICE baselines with the help of extra simulator. These results not only empirically validate the effectiveness of our algorithm, but also point out that we can learn important information from imperfect simulated environments to help in improving RL policies. We also observe that as the Sim2Sim gap increases the performance of our algorithm tends to decline. This means that with increasing Sim2Sim gap the amount of relevant information that can be learned from the simulator diminishes. We observe that this decline actually becomes significant when the Sim2Sim gap breaches the 60\% threshold. 

\vspace{-6pt}
\subsection{Real-world performance validation on Kinova Robotic Arm}
\vspace{-3pt}
\textbf{Experimental Setting:} We demonstrate the effectiveness of our $\beta$-DICE algorithm for a sim2real validation task on a Kinova robotic arm. We first collect offline data by asking users to move the arm from one-position to another via RC controllers. Our data collection ensures sufficient coverage of the robotic arm's task space. We then use this offline data along with our in-house gazebo based simulator to experimentally validate the real-world performance of a PID controller using $\beta$-DICE that moves our robot from a given initial location to any desired location.\\
\textbf{Results and Observations:} Our results along with different baselines are averaged over 10 different locations are tabulated in Table \ref{tab:kinova}. We observe that $\beta$-DICE is able to outperform state-of-the-art showing an improvement of 60\% over the nearest baseline. There are two key conclusions from all of our experiments. One, although $\beta$-DICE outperforms state-of-the art baselines in off-policy evaluation. We observe that the performance drops when the gap between the target policy and behavior policy increases. Two, prediction error decreases as the gap between the training and test environments increases, as the transferable information between the two environments decreases. 
\vspace{-5pt}

\vspace{-5pt}
\section{Limitations}
\vspace{-5pt}
We present the limitations of our work that we wish to address in future work. (1) Our algorithm expects high quality data with sufficient coverage of the state-action space. Identifying the confidence interval of our estimator $w$ will not only ensure sample efficient evaluation, but also help us in designing robust offline reinforcement learning algorithm. (2) Similar to DICE class of min-max optimization, our algorithm also suffers from high variance in their performance. Efforts are required to reduce this variance. 
\vspace{-5pt}
\section{Conclusion and Future Work} \label{sec:conclusion}
\vspace{-5pt}
We derive a novel MIS estimator that is able to evaluate real world performance of a robot using offline data and an imperfect robot simulator. We then develop sample complexity bounds, and empirically validate our approach on diverse Sim2Sim environments and Sim2Real environment like KinovaGen3 robot. For future work, we wish to utilize this framework of off-environment evaluation to learn optimal robot policies using simulation and a limited amount of real-world offline data. 

\section{Acknowledgements}
The authors thank Neeloy Chakroborty and Shuijing Liu for their valuable suggestions on this paper draft. 
This work was supported in part by ZJU-UIUC Joint Research Center Project No. DREMES 202003, funded by Zhejiang University.
Additionally, Nan Jiang would also like to acknowledge funding support from NSF IIS-2112471 and NSF CAREER IIS-2141781.

\newpage

\bibliography{example}
\onecolumn
\section{Appendix}
\subsection{Case Studies}\label{sec:case_studies}
As a common issue in MIS, the general estimators are usually difficult to optimize due to the minimax form. One solution is to choose the discriminator class ($\mathcal{Q}$ in our case) to be an RKHS, which often leads to a closed-form solution to the inner max and reduces the minimax optimization to a single minimization problem \cite{liu2018breaking, uehara2019minimax, voloshin2021minimax}. Below we show that this is also the case for our estimator, and provide the closed-form expression for the inner maximization when $\mathcal{Q}$ is an RKHS. 
\begin{lemma}\label{le:RKHS}
Let $\langle ., . \rangle_{\mathcal{H}_K}$ be the inner-product of $\mathcal{H}_K$ which satisfies the Reproducible Kernel Hilbert Space (RKHS) property. When the function space $\mathcal{Q} = \{q: \mathcal{\mathcal{S} \times \mathcal{A}} \rightarrow \mathbb{R};\, \langle q, q \rangle_{\mathcal{H}_K} \leq 1 \}$, the term $\max_{q \in \mathcal{Q}} L_w(w, \beta, q)^2$ has the following closed-form expression: 
\begin{align*}
    &\mathbb{E}_{\substack{(s, a, s') \sim \mu \\
    (\tilde{s}, \tilde{a}, \tilde{s}') \sim \mu}}[w(s, a) \cdot w(\tilde{s}, \tilde{a}) \cdot \beta(s, a) \cdot \beta(\tilde{s}, \tilde{a}) \cdot \Large(K((s, a), (\tilde{s}, \tilde{a}))-2 \gamma \mathbb{E}_{a' \sim \pi(\cdot|s')}[K((s' , a'), (\tilde{s}, \tilde{a}))]\\
    &+\gamma^2 \mathbb{E}_{\substack{a' \sim \pi(.
    s') \\ \tilde{a}' \sim \pi(.
    \tilde{s}')}}[K((s', a'), (\tilde{s}', \tilde{a}'))]\Large)]- 2(1-\gamma)\mathbb{E}_{\substack{(s, a, s') \sim \mu \\
    \tilde{s} \sim d_0, \tilde{a} \sim \pi(\cdot|\tilde{s})}}[w(s, a)\cdot \beta(s, a) \cdot (K((s, a), (\tilde{s}, \tilde{a}))  \\&-\gamma \mathbb{E}_{a' \sim \pi(.
    s')} [K((s', a'), (\tilde{s}, \tilde{a}))]]  + (1-\gamma)^2\mathbb{E}_{\substack{s\sim d_0,a \sim \pi(\cdot|s) \\ \tilde{s} \sim d_0, \tilde{a} \sim \pi(\cdot | \tilde{s})}}[K((s, a), (\tilde{s}, \tilde{a}))].
\end{align*}
\end{lemma}

Furthermore, when we use linear functions to approximate both $w$ and $q$, the final estimator has a  closed-form solution 
\begin{lemma}\label{le:linear_features}
Consider linear parameterization $w(s,a)  = \phi(s, a)^T \alpha$, where $\phi \in \mathbb{R}^{d}$ is a feature map in $\mathbb{R}^d$ and $\alpha$ is the linear coefficients. Similarly let $q(s, a) = \Psi(s, a)^T \zeta$ where $\Psi \in \mathbb{R}^{d}$. Then, assuming that we have an estimate of $\frac{d^\pi_{P_{tr}}}{\mu}$ as $\hat\beta$, we can empirically estimate $\hat w$ using Equation~\ref{eq:optimizer}, which has a closed-form expression $\hat w(s,a) = \phi(s,a)^T \hat \alpha$, where
\begin{align}
\begin{split}
    \hat\alpha &= (\mathbb{E}_{n, (s, a, s') \sim \mu}[(\Psi(s, a) - \gamma \Psi(s', \pi)) \cdot \phi(s,a)^T \cdot\hat\beta(s,a)])^{-1} 
    (1 - \gamma) \mathbb{E}_{n, s \sim d_0}[\Psi(s, \pi)]
    \end{split}
\end{align}
provided that the matrix being inverted is non-singular. Here, $\mathbb{E}_n$ is the empirical expectation using $n$-samples. 
\end{lemma}
Detailed proof for these Lemma can be found in section \ref{sec:proof_RKHS} and \ref{sec:proof_linear_features} respectively. 

\subsection{Q-Function Estimator}\label{sec:q-estimator}
In this section, we show an extension of our idea that can approximate the Q-function in the target environment. 
Similar to we did in the previous section, we now consider the OPE error of a candidate function $q$, that is, $|(1-\gamma)\mathbb{E}_{s \sim d_0}[q(s, \pi)] - J(\pi)|$, under the assumption that $w_{P_{te}/P_{tr}} \in conv(\mathcal{W})$:
\begin{align}\label{eq:derivation-q}
\begin{split}
& |(1-\gamma)\mathbb{E}_{s \sim d_0}[q(s, \pi)] - J_{P_{te}}(\pi)| 
= |\mathbb{E}_{\substack{(s,a) \sim d_{P_{te}}^\pi,\\ r \sim R(s,a), s'\sim P(s,a)}}[q(s,a)  - \gamma q(s', \pi)] - \mathbb{E}_{\substack{(s, a) \sim d^\pi_{P_{tr}} \\r \sim R(s, a) }}[W_{P_{te}/P_{tr}} \cdot r]| \\
= &~ |\mathbb{E}_{\substack{(s,a) \sim \mu,\\ r \sim R(s,a), s'\sim P(s,a)}}[W_{P_{te}/P_{tr}} \cdot \beta \cdot (q(s,a) - \gamma q(s', \pi))]- \mathbb{E}_{\substack{(s, a) \sim d^\pi_{P_{tr}} \\r \sim R(s, a) }}[W_{P_{te}/P_{tr}} \cdot r]| \\
\le &~ \sup_{w\in\mathcal{W}} |\mathbb{E}_{\substack{(s,a) \sim \mu,\\ r \sim R(s,a), s'\sim P(s,a)}}[w \cdot \beta \cdot (q(s,a) - \gamma q(s', \pi) )]-\mathbb{E}_{\substack{(s, a) \sim d^\pi_{P_{tr}} \\r \sim R(s, a) }}[w \cdot r]|  \\
=: & \sup_{w \in \mathcal{W}} L_q(w, \beta, q).
\end{split}
\end{align}
The inequality step uses the assumption that $w_{P_{te}/P_{tr}} \in conv(\mathcal{W})$, and the final expression is a valid upper bound on the error of using $q$ for estimating $J_{P_{te}}(\pi)$. It is also easy to see that the bound is tight because $q=Q_{P_{te}}^\pi$ satisfies the Bellman equation on all state-action pairs, and hence $L_q(w, \beta, Q_{P_{te}}^\pi) \equiv 0$.

Using this derivation, we propose the following estimator which will estimate $Q^\pi_{P_{te}}$. 
\begin{align}\label{eq:q_approx}
    Q^\pi_{P_{te}} \approx \hat{q} := \arg \min_{q \in \mathcal{Q}} \max_{w \in \mathcal{W}} L_q(w, \beta, q).
\end{align}

Below we provide the results that parallel Lemmas~\ref{le:RKHS} and \ref{le:linear_features} for the Q-function estimator. 
\begin{lemma}\label{le:RKHS-q}
Let $\langle ., . \rangle_{\mathcal{H}_K}$ be the inner-product of $\mathcal{H}_K$ which satisfies the Reproducible Kernel Hilbert Space (RKHS) property. When the function space $\mathcal{W} = \{w: \mathcal{S} \times \mathcal{A} \rightarrow \mathbb{R} | \langle w, w \rangle_{\mathcal{H}_K} \leq 1 \}$. The term $\max_{w \in \mathcal{W}} L_q(w, \beta, q)^2$ has a closed form expression.
\end{lemma}
We defer the detailed expression and its proof to Appendix~\ref{app:RKHS-q}.
\begin{lemma}\label{le:linear_features_1}
Let $w  = \phi(s, a)^T \alpha$ where $\phi \in \mathbb{R}^{d}$ is some basis function. Let $q(s, a) = \Psi(s, a)^T \zeta$, where $\Psi(s, a) \in \mathbb{R}^d$. Then, assuming that we have an estimate of $\frac{d^\pi_{P_{tr}}}{\mu}$ as $\hat\beta$, we can empirically estimate $\hat q$ using uniqueness condition similar to Equation~\ref{eq:q_approx}, which has a closed-form expression $\hat w(s,a) = \Psi(s,a)^T \hat{\zeta}$, where
\begin{align}
\begin{split}
    \hat{\zeta} &= (\mathbb{E}_{n, \mu}^{\pi}[{\hat\beta
    } \cdot (\Phi(s, a) \Psi(s, a)^T   - \gamma \Phi(s, a) \Psi(s', \pi)]))^{-1} \mathbb{E}_{n, (s, a) \sim d^\pi_{P_{tr}},  r \sim R(s, a))}[\Phi(s, a) \cdot r] 
    \end{split}
\end{align}
where, $\mathbb{E}_n$ is the empirical expectation calculated over n-samples and assuming that the provided matrix is non-singular. 
\end{lemma}
\begin{theorem}\label{thm:finite-sample-2}
Let $\hat \beta$ be our estimation of $\beta$ using \cite{DBLP:journals/corr/abs-0809-0853}. We utilize this $\hat{\beta}$ to further optimize for $\hat{w}_n$ (equation \ref{eq:optimizer}) using n samples. In both cases, $\mathbb{E}_{(s,a) \sim d_{P_{tr}}^\pi}[\cdot]$ is also approximated with $n$ samples from the simulator $P_{tr}$. 
Then, under Assumptions~\ref{ass: bounds} and \ref{ass: existence} along with the additional assumption that $Q^\pi_{P_{te}} \in C(\mathcal{Q})$ with probability at least $1-\delta$, We can guarantee the OPE error for $\hat{q}_n$ which was optimized using equation \ref{eq:q_approx} on n samples. 
\begin{align*}
    &|(1-\gamma)\mathbb{E}_{d_0} [\hat{q}_n(s, \pi)] - J_P(\pi)|\leq \\
    & \min_{q \in \mathcal{Q}} \max_{w \in \mathcal{W}} L_q(w, \beta, q) +  4 \mathcal{R}_n(\mathcal{W}, \mathcal{Q}) + 2C_{\mathcal{W}} \frac{R_{max}}{1-\gamma}\sqrt{\frac{\log  (\frac{2}{\delta})}{2n}}   \\
    &+ C_{\mathcal{W}} \frac{R_{max}}{1 - \gamma}\cdot \tilde{O}\left(  \sqrt{  \|\frac{d^\pi_{P_{tr}}}{\mu} \|_\infty \left(4\mathbb{E} \mathcal{R}_n(\mathcal{F}) + C_{\mathcal{F}}\sqrt{\frac{2 \log(\frac{2}{\delta})}{n}}\right)}\right)
\end{align*} 
where $\mathcal{R}_n(\mathcal{F}), \mathcal{R}_n(\mathcal{W}, \mathcal{Q})$ are the Radamacher complexities of function classes $\{(x, y) \rightarrow f(x) - \log(f(y)) : f \in\mathcal{F}\}$  and 
$\{(s, a, s') \rightarrow (w(s, a) \cdot \frac{d^\pi_{P'}(s, a)}{\mu(s, a)} \cdot (q(s, a) - \gamma q(s', \pi)) : w\in\mathcal{W}, q\in\mathcal{Q}\} $, respectively, 
$\|d^\pi_{P'}/\mu\|_\infty := \max_{s,a} d^\pi_{P'}(s,a) / \mu(s,a)$ measures the distribution shift between $d^\pi_{P'}$ and $\mu$, 
and $\tilde O(\cdot)$ is the big-Oh notation suppressing logarithmic factors.  Under the assumption $w^{\pi}_{P_{tr}/P_{te}} \in C(\mathcal{W})$, 
\end{theorem}
\subsection{Derivation for $\beta$-GradientDICE}\label{sec:beta_gradient_dice}
We will show a demonstration on finite state-action space. The following identity holds true for $\tau_* = \frac{d^{\pi}_{P_{te}}}{d^{\pi}_{P_{tr}}}$. Let us assume that we have the diagonal matrix $D$ with diagonal elements being $d^{\pi}_{P_{tr}}$. The following identity holds true. 
\begin{align}
D\tau_* = \mathcal{T}\tau_*
\end{align}
Where, $d_0(s, a) = d_0(s)\pi(a | s)$ and $\mathcal{T}$ is the reverse bellman operator
\begin{align*}
    \mathcal{T}y = (1 - \gamma)d_0(s, a) + \gamma P_{\pi}^T D y
\end{align*}
Where, $P_\pi((s, a), (s', a')) = P_{te}(s' | s, a)\pi(a' | s')$
To estimate $\tau$, we can simply run the following optimization
\begin{align*}
    \tau := \arg \min_{\tau:S \times A \rightarrow \mathbb{R}}|D\tau - \mathcal{T} \tau |_{D^{-1}}^2 + \frac{\lambda}{2} ((d^{\pi}_{P_{tr}})^T \tau - 1)
\end{align*}
Here, $|y|_{\Sigma}^2 = y^T \Sigma y$. The optimization above can be simplified in form of expectation over $d^{\pi}_{P_{tr}}$. 
\begin{align*}
    \mathbb{E}_{(s, a) \sim d^{\pi}_{P_{tr}}}[(\frac{\delta(s, a)}{d^{\pi}_{tr}(s, a)})^2] + \frac{\lambda}{2}((d^{\pi}_{P_{tr}})^T \tau - 1)
\end{align*}
With, $\delta(s, a) = D\tau - \mathcal{T}\tau $, We can now apply Fenchel Conjugate principle to get the following
\begin{align*}
    \max_{f: S \times A \rightarrow \mathbb{R}}\mathbb{E}_{(s, a) \sim d^{\pi}_{P_{tr}}}[\frac{\delta(s, a)}{d^\pi_{P_{tr}}}f(s, a) - \frac{1}{2}f(s, a)^2] + \max_{\eta \in \mathbb{R}} (\mathbb{E}_{d^{\pi}_{P_{tr}}}[\eta \tau(s, a) - \eta] - \frac{\eta^2}{2})
\end{align*}
If we simplify the above optimization, we get the following form
\begin{align*}
        &\frac{d^{\pi}_{P_{te}}}{d^{\pi}_{P_{te}}} := \arg \min_{\tau: S \times A \rightarrow \mathbb{R}} \max_{f: S \times A \rightarrow \mathbb{R}, \eta \in \mathbb{R}} L(\tau, \eta, f) \\
        & = (1 - \gamma) \mathbb{E}_{s_0 \sim d_0, a_0 \sim \pi(\cdot | s_0)}[f(s_0, a_0)]+ \gamma \mathbb{E}_{\substack{(s, a) \sim d^{\pi}_{P_{tr}}\\s' \sim P_{te}(\cdot | s, a), a' \sim \pi(\cdot | s')}}[\tau(s, a) f(s', a')] \\
        &- \mathbb{E}_{(s, a) \sim d^{\pi}_{P_{tr}}}[\tau(s, a) f(s, a)] - \frac{1}{2}\mathbb{E}_{(s, a) \sim d^{\pi}_{P_{tr}}}[f(s, a)^2] + \lambda \mathbb{E}_{(s, a) \sim d^{\pi}_{P_{tr}}}[\eta \tau(s, a) - \eta^2/2].
\end{align*}
While we don't have samples from $(s, a, s') \sim d^{\pi}_{P_{tr}}$. We can simply re-weight the term $\mathbb{E}_{\substack{(s, a) \sim d^{\pi}_{P_{tr}}\\s' \sim P_{te}(\cdot | s, a), a' \sim \pi(\cdot | s')}}[\tau(s, a) f(s', a')]$ with $\beta(s, a) = \frac{d^{\pi}_{P_{tr}}}{\mu}$. This completes the derivation of $\beta$-GradientDICE. 
\begin{align*}
        &\frac{d^{\pi}_{P_{te}}}{d^{\pi}_{P_{te}}} := \arg \min_{\tau: S \times A \rightarrow \mathbb{R}} \max_{f: S \times A \rightarrow \mathbb{R}, \eta \in \mathbb{R}} L(\tau, \eta, f) \\
        & = (1 - \gamma) \mathbb{E}_{s_0 \sim d_0, a_0 \sim \pi(\cdot | s_0)}[f(s_0, a_0)]+ \gamma \mathbb{E}_{(s, a, s') \sim \mu, a' \sim \pi(\cdot | s')}[\beta(s, a) \tau(s, a) f(s', a')] \\
        &- \mathbb{E}_{(s, a) \sim d^{\pi}_{P_{tr}}}[\tau(s, a) f(s, a)] - \frac{1}{2}\mathbb{E}_{(s, a) \sim d^{\pi}_{P_{tr}}}[f(s, a)^2] + \lambda \mathbb{E}_{(s, a) \sim d^{\pi}_{P_{tr}}}[\eta \tau(s, a) - \eta^2/2].
\end{align*}
\subsection{Proof of Lemma \ref{le:RKHS}}\label{sec:proof_RKHS}
Since $\mathcal{Q}$ belongs to the RKHS space. We can use the reproducible property of RKHS to re-write the optimization in the following form. 
\begin{align}
\begin{split}
&L_w(w, \beta, q)^2=(\mathbb{E}_{(s, a) \sim \mu, s' \sim P_{te}(s, a)}[w(s, a)\cdot \beta(s, a) \cdot (q(s,a) -\gamma q(s', \pi))] - (1-\gamma) \mathbb{E}_{s \sim d_0}[q(s, \pi)])^2\\
    &=(\mathbb{E}_{(s, a) \sim \mu, s' \sim P_{te}(s, a)}[w(s, a)\cdot \beta(s, a) \cdot (\langle q, K((s, a), .), \cdot \rangle_{\mathcal{H}_K} -\gamma \mathbb{E}_{a' \sim \pi(.
    s')} [\langle q, K((s', a'), .), \cdot \rangle_{\mathcal{H}_K}]] \\
    &\quad \quad \quad  -(1-\gamma) \mathbb{E}_{s \sim d_0, a \sim \pi(. | s)}[\langle q, K((s, a), .), \cdot \rangle_{\mathcal{H}_K}]))^2\\
    &= \max_{q \in \mathcal{Q}}\langle q, q^* \rangle_{\mathcal{H}_K}^2 
    \end{split}
\end{align}
Where, 
\begin{align}
\begin{split}
    q^*(\cdot) &= \mathbb{E}_{\mu}[w(s, a)\cdot \beta(s, a) \cdot (K((s, a), .) -\gamma \mathbb{E}_{a' \sim \pi(.
    s')} [K((s', a'), .)]] - (1-\gamma) \mathbb{E}_{s \sim d_0, a \sim \pi(. | s)}[K((s, a), .)])
    \end{split}
\end{align}

We go from first line to the second line by exploiting the linear properties of the RKHS function space. Given the constraint that $\mathcal{Q} = \{q: \mathcal{\mathcal{S} \times \mathcal{A}} \rightarrow \mathbb{R};\, \langle q, q \rangle_{\mathcal{H}_K} \leq 1 \}$ we can maximise $\max_q L(w, \beta, q)^2$ using Cauchy-Shwartz inequality 
\begin{align*}
    &\max_{q} L_w(w, \beta, q)^2 =  \langle q^*, q^* \rangle_{\mathcal{H}_K}^2\\
    & = \mathbb{E}_{\substack{(s, a, s') \sim \mu \\
    (\tilde{s}, \tilde{a}, \tilde{s}') \sim \mu}}[w(s, a) \cdot w(\tilde{s}, \tilde{a}) \cdot \beta(s, a) \cdot \beta(\tilde{s}, \tilde{a}) \cdot  (K((s, a), (\tilde{s}, \tilde{a})) -2 \gamma \mathbb{E}_{a' \sim \pi(\cdot|s')}[K((s' , a'), (\tilde{s}, \tilde{a}))]\\
    &+\gamma^2 \mathbb{E}_{\substack{a' \sim \pi(.
    s') \\ \tilde{a}' \sim \pi(.
    \tilde{s}')}}[K((s', a'), (\tilde{s}', \tilde{a}'))])]- 2(1-\gamma)\mathbb{E}_{\substack{(s, a, s') \sim \mu \\
    \tilde{s} \sim d_0, \tilde{a} \sim \pi(\cdot|\tilde{s})}}[w(s, a)\cdot \beta(s, a) \cdot (K((s, a), (\tilde{s}, \tilde{a}))  \\&-\gamma \mathbb{E}_{a' \sim \pi(.
    s')} [K((s', a'), (\tilde{s}, \tilde{a}))]]  + (1-\gamma)^2\mathbb{E}_{\substack{s\sim d_0,a \sim \pi(\cdot|s) \\ \tilde{s} \sim d_0, \tilde{a} \sim \pi(\cdot | \tilde{s})}}[K((s, a), (\tilde{s}, \tilde{a}))]\\
\end{align*}
This completes the proof.

\subsection{Proof of Lemma \ref{le:linear_features}}\label{sec:proof_linear_features}
Substituting the functional forms for $q(s, a) = \Psi(s, a)^T\zeta$ and $w(s, a) = \phi(s, a)^T\alpha$ we get the following expression for $L_{n, w}(w, \hat{\beta}, q)$. Where, $\hat{\beta}$ is an estimate of $\frac{{d}^\pi_{P_{tr}}}{\mu}$
\begin{align*}
     L_{n, w}(w, \hat{\beta}, q) &=\mathbb{E}_{n, \mu }[\phi(s,a)^T \alpha \cdot \hat{\beta}(s, a) \cdot (\Psi(s, a) - \gamma \Psi(s', \pi))^T \zeta)] - (1 - \gamma) \mathbb{E}_{n, d_0}[\Psi(s, \pi)^T \zeta]
\end{align*}
Using the uniqueness condition we derived in equation \ref{eq:uniqueness}, we can go about finding the value of $\alpha$ by equating $L(w, \hat{\beta}, q)$ to zero. 
\begin{align*}
    &\mathbb{E}_{n, \mu}[\phi(s,a)^T \alpha \cdot \hat{\beta}(s, a) \cdot (\Psi(s, a) - \gamma \Psi(s', \pi))^T \zeta)] - (1 - \gamma) \mathbb{E}_{n, d_0}[\Psi(s, \pi)^T \zeta]  = 0 \\
     &\alpha^T\mathbb{E}_{n, \mu}[\phi(s,a) \cdot \hat{\beta}(s, a) \cdot (\Psi(s, a) - \gamma \Psi(s', \pi))^T)]\zeta  = (1 - \gamma) \mathbb{E}_{n, d_0}[\Psi(s, \pi)^T] \zeta
\end{align*}
Since the loss is linear in $\zeta$, we can solve for $\alpha$ using the matrix inversion operation. 
     \begin{align*}
     \hat{\alpha} = \LARGE(\mathbb{E}_{n, \mu}[(\Psi(s, a) - \gamma \Psi(s', \pi)) \cdot \phi(s,a)^T \cdot \hat{\beta}]\LARGE)^{-1} (1 - \gamma) \mathbb{E}_{n, d_0}[\Psi(s, \pi)]
\end{align*}
This completes the proof. 
\subsection{Proof of Lemma \ref{le:RKHS-q}} \label{app:RKHS-q}
Consider the loss function $\sup_{w \in \mathcal{W}} L_q(w, \beta, q)^2$. Since $\mathcal{W}$ is in RKHS space. Using reproducible property of RKHS space we can re-write this maximization as follows, 
\begin{align*}
&\max_{w \in \mathcal{W}}L_q(w, \beta, q)^2 = \max_{w \in \mathcal{W}}(\mathbb{E}_{(s,a) \sim \mu, s' \sim P(s,a), r \sim R(s, a)}[w(s, a) \cdot \beta(s, a) \cdot (q(s,a) -\gamma q(s', \pi))] - \mathbb{E}_{(s, a) \sim d^\pi_{P_{tr}},r \sim R(s, a)}[w(s, a) \cdot r])^2  \\
& \max_{w \in \mathcal{W}} (\mathbb{E}_{(s,a) \sim \mu, s' \sim P(s,a), r \sim R(s, a)}[\langle w, K(s, a), \cdot \rangle_{\mathcal{H}_K} \cdot \beta(s, a) \cdot (q(s,a) -\gamma q(s', \pi))] - \mathbb{E}_{(s, a) \sim d^\pi_{P_{tr}},r \sim R(s, a)}[\langle w, K(s, a), \cdot \rangle_{\mathcal{H}_K} \cdot r])^2  \\
&\max_{w \in \mathcal{W}} \langle w, \mathbb{E}_{(s,a) \sim \mu, s' \sim P(s,a), r \sim R(s, a)}[K((s, a), \cdot) \cdot \beta(s, a) \cdot (q(s,a) -\gamma q(s', \pi))] - \mathbb{E}_{(s, a) \sim d^\pi_{P_{tr}},r \sim R(s, a)}[K((s, a), \cdot) \cdot r]\rangle_{\mathcal{H}_K})^2 \\
&\max_{w \in \mathcal{W}} \langle w, w^* \rangle_{\mathcal{H}_K}^2 = \langle w^*, w^* \rangle_{\mathcal{H}_K}^2
\end{align*}
Where, we use the linear properties of RKHS spaces and then followed by using Cauchy-Shwartz inequality, to compute the maximization. Where, $w^*$ has the following expression. 
\begin{align*}
    w^*(\cdot) = \mathbb{E}_{(s,a) \sim \mu, s' \sim P(s,a), r \sim R(s, a)}[K((s, a), \cdot) \cdot \beta(s, a) \cdot (q(s,a) -\gamma q(s', \pi))] - \mathbb{E}_{(s, a) \sim d^\pi_{P_{tr}},r \sim R(s, a)}[K((s, a), \cdot) \cdot r]
\end{align*}
The maximization expression thus takes the following form
\begin{align*}
   & \langle w^*, w^* \rangle_{\mathcal{H}_K}^2 = \mathbb{E}_{\substack{(s,a) \sim \mu, s' \sim P(s,a), r \sim R(s, a) \\ (\tilde{s},\tilde{a}) \sim \mu, \tilde{s}' \sim P(s,a), r \sim R(s, a)}}[K((s, a), (\tilde{s}, \tilde{a})) \cdot \beta(s, a) \cdot \beta (\tilde{s}, \tilde{a}) \cdot \Delta(q, s, a, s') \cdot \Delta(q, \tilde{s}, \tilde{a}, \tilde{s}')]   \\
    &-2 \mathbb{E}_{\substack{(s,a) \sim \mu, s' \sim P(s,a) \\ (\tilde{s},\tilde{a}) \sim d^\pi_{P_{tr}}, \tilde{r} \sim R(s, a)}, }[K((s, a), (\tilde{s}, \tilde{a})) \cdot \beta(s, a) \cdot \Delta(q, s, a, s') \cdot r] + \mathbb{E}_{\substack{(s, a) \sim d^\pi_{P_{tr}}, r \sim R(s, a) \\ (\tilde{s}, \tilde{a}) \sim d^\pi_{P_{tr}}, \tilde{r} \sim R(s, a)}}[K((s, a), (\tilde{s}, \tilde{a})) \cdot r \cdot \tilde{r}]
\end{align*}
Where, $\Delta(q, s, a, s') = q(s, a) - \gamma q(s', \pi)$.\\
This completes the proof.
\subsection{Proof of Lemma \ref{le:linear_features_1}}
Substituting the functional forms of $q(s, a) = \Psi(s, a)^T \zeta$, $w(s, a) = \phi(s, a)^T \alpha$. Also substituting the estimate for $\frac{d^\pi_{P_{tr}}}{\mu}$ as $\hat{\beta}$. We get the following expression
\begin{align*}
&L_{q, n}(w, \hat{\beta}, q) = \\
&|\mathbb{E}_{n, (s,a) \sim \mu, s' \sim P(s,a), r \sim R(s, a)}[ \phi(s, a)^T \alpha \cdot \hat{\beta}(s, a)\cdot (\Psi(s, a)^T \zeta -\gamma \Psi(s', \pi)^T \zeta)] - \mathbb{E}_{n, (s, a) \sim d^\pi_{P_{tr}},r \sim R(s, a)}[\phi(s, a)^T \alpha \cdot r]| \\
&= 0
\end{align*}
Where, the equality comes from the uniqueness condition similar to equation \ref{eq:uniqueness}
\begin{align*}
    \alpha^T\mathbb{E}_{n, (s,a) \sim \mu, s' \sim P(s,a), r \sim R(s, a)}[\phi(s, a) \cdot \hat{\beta}(s, a) \cdot (\Psi(s, a) - \gamma \Psi(s', \pi)) ^T]\zeta = \alpha^T\mathbb{E}_{n, (s, a) \sim d^\pi_{P_{tr}},r \sim R(s, a)}[\phi(s, a) \cdot r]
\end{align*}
Since the equations above are linear in $\alpha$. So it suffices to show that the optimal solution can be reached if $\beta$ is approximated as follows, 
\begin{align}
    \hat{\zeta} = (\mathbb{E}_{n, (s,a) \sim \mu, s' \sim P(s,a), r \sim R(s, a)}[\phi(s, a) \cdot \hat{\beta}(s, a) \cdot (\Psi(s, a) - \gamma \Psi(s', \pi)) ^T])^{-1} \cdot \mathbb{E}_{n, (s, a) \sim d^\pi_{P_{tr}},r \sim R(s, a)}[\phi(s, a) \cdot r]
\end{align}
Where, $\mathbb{E}_{n, \cdot}$ denotes the empirical approximation of the expectation. This completes the proof. 

\subsection{Proof of Theorem \ref{thm:finite-sample}}
To prove this theorem, we will first require a Lemma that we need to prove first. This is as follows, 
\begin{lemma} \label{le: estimation}
Under Assumptions \ref{ass: bounds} and \ref{ass: existence}, suppose we use $n$ samples each from distribution $P$ and $Q$ to empirically estimate the ratio of $\frac{P}{Q}$ using equation \ref{eq:beta_optimizer}. The estimation error can be bounded with probability at least $1- \delta$ as follows:
\begin{align}\label{eq: upperbound}
\begin{split}
  \left \| \hat{f}_n - \frac{P}{Q} \right\|_\infty^2 \leq \tilde{O} \left( \|\frac{P}{Q}\|_\infty \left(4\mathbb{E} \mathcal{R}_n(\mathcal{F}) + \sqrt{\frac{2 \log(\frac{1}{\delta})}{n}}\right)\right)  
\end{split}
\end{align}
\end{lemma}
\begin{proof}
Since, equation \ref{eq:beta_optimizer} is optimized using empirical samples it is an Empirical Risk Minimization (ERM) algorithm. We denote the original loss with respect to a function $f \in \mathcal{F} $ as $L(f)$.  Using familiar result from learning theory (Corollary 6.1 ~\cite{DBLP:journals/corr/Hajek}) with probability at-least $1-\delta$
\begin{align}\label{eq: radaapprox}
    L(\hat{f}_n) - L(\frac{P}{Q}) \leq 4 \mathbb{E} \mathcal{R}_n(\mathcal{F}) + C_{\mathcal{F}}\sqrt{\frac{2 \log(\frac{1}{\delta})}{n}} 
\end{align}
With probability at least $1-\delta$. Where, $\mathcal{R}_n(\mathcal{F})$ is the Radamacher complexity of the function class 
\begin{align}
    \{(p, q) \rightarrow  f(q) - \log(f(p): f \in \mathcal{F}\}
\end{align}
Now, let's turn our attention to the left hand side. Before we end up doing that let's define the estimation error $\Bar{e}_n(x) = \hat{f}_n(x) - \frac{P(x)}{Q(x)}$. Thus, we can re-write the left hand side in terms of $\Bar{e}$
\begin{align}
\begin{split}
    &L(\hat{f}_n) - L(\frac{P}{Q})  = L(\frac{P}{Q} + \Bar{e}_n) - L(\frac{P}{Q})\\
    &= \sum_{x \in \Omega} Q(x) \bar{e}_n(x) 
    -\sum_{x \in \Omega} P(x) log(\frac{\Bar{e}_n(x) + \frac{P(x)}{Q(x)}}{\frac{P(x)}{Q(x)}}) \\
    &= \sum_{x \in \Omega} Q(x) (\bar{e}_n(x) - \frac{P(x)}{Q(x)} \log(1 + \frac{\bar{e}_n(x)}{\frac{P(x)}{Q(x)}})) 
    \end{split}
\end{align}
Assuming that $n$ is sufficiently large such that $|\frac{\Bar{e}_n}{g^*}| \leq 1$. We can now use second order Taylor approximation for $\log (1 + x)$ for $|x| < 1$
\begin{align}\label{eq:taylorapprox}
\begin{split}
     L(\hat{f}_n) - L(\frac{P}{Q})  &= \sum_{x \in \Omega} Q(x) (\Bar{e}_n(x) \\
     &- \frac{P(x)}{Q(x)}\cdot \left( \frac{\Bar{e}_n(x)}{\frac{P(x)}{Q(x)}} - \frac{1}{2} (\frac{\Bar{e}_n(x)}{\frac{P(x)}{Q(x)}})^2) \right)
     \\
     &= \sum_{x \in \Omega} Q(x)\frac{1}{2}(\frac{\Bar{e}_n(x)^2}{\frac{P(x)}{Q(x)}})
     \end{split}
\end{align}
Combining equations \ref{eq: radaapprox} with the simplified LHS above, we can bound the error with probability at least $1-\delta$ that, 

\begin{align}
\begin{split}
\sum_{x \in \Omega} &Q(x)\frac{1}{2}(\frac{\Bar{e}_n(x)^2}{\frac{P(x)}{Q(x)}}) \leq
4 \mathbb{E} \mathcal{R}_n(\mathcal{F}) + C_{\mathcal{F}}\sqrt{\frac{2 \log(\frac{1}{\delta})}{n}} 
\end{split}
\end{align}
Under assumption \ref{ass: bounds} and \ref{ass: existence} $\exists \tilde{x} \in \Omega$ such that $|\bar{e}_n(\tilde{x})| =  \| \hat{f}_n - \frac{P}{Q}\|_\infty$. Thus, the equation above can be re-written as 
\begin{align}\label{eq:beta_error}
\begin{split}
    &\frac{1}{K }\| \Bar{e}_n \|_\infty^2 \leq  2 \frac{P(\tilde{x})}{Q(\tilde{x})} \left(4\mathbb{E} \mathcal{R}_n(\mathcal{F}) + C_{\mathcal{F}}\sqrt{\frac{2 \log(\frac{1}{\delta})}{n}}\right)  \\
    &\| \Bar{e}_n \|_\infty^2 \leq 2K \cdot \|\frac{P}{Q} \|_\infty \left(4\mathbb{E} \mathcal{R}_n(\mathcal{F}) + C_{\mathcal{F}}\sqrt{\frac{2 \log(\frac{1}{\delta})}{n}}\right)
    \end{split}
\end{align}
Where $Q(\tilde{x}) = \frac{1}{K}$. The last inequality comes from the fact that $\frac{P(\tilde{x})}{Q(\tilde{x})} \leq \sup_{x \in \Omega} \frac{P(x)}{Q(x)} = \|\frac{P}{Q} \|_\infty$. 
This completes the proof. 
\end{proof}
Using equation \ref{eq:derivation} we can upper bound the performance of our estimator as follows, 
      \begin{align}
     \begin{split}
    &|\mathbb{E}_{(s, a) \sim d^\pi_{\mathcal{P}}, r \sim R(s, a)} [\hat{w}_n \cdot r] - J_P(\pi)| \leq  \max_{q \in \mathcal{Q}} |L_w(\hat{w}_n,  \frac{d^\pi_{P}}{\mu}, q) | \\
    & \hat{w}_n = \arg \min_{w \in \mathcal{W}} \max_{q \in \mathcal{Q}} L_{n, w} (w, \hat{\beta}, q) 
    \end{split}
 \end{align}
 We also approximate $\frac{d^\pi_{P_{tr}}}{\mu} \sim \hat{\beta}$. This can be written as follows, 
 \begin{align}
   \hat{\beta} = \arg \max_{f \in \mathcal{F}} \frac{1}{n}\sum_{i} \ln f(x_i) - \frac{1}{m}\sum_{j}f(\tilde{x}_j) + \frac{\lambda}{2} I(f)^2,
\end{align}
where $I(f)$ is some regularization function to improve the statistical and computational stability of learning. We can the simplify the RHS of this upper-bound using the following simplification. 
 \begin{align}
 \begin{split}
     &|\mathbb{E}_{(s, a) \sim d^\pi_{\mathcal{P'}}, r \sim R(s, a)} [\hat{w}_n \cdot r] - J_P(\pi)|  \leq  \max_{q \in \mathcal{Q}} |L_w(\hat{w}_n, \frac{d^\pi_{P'}}{\mu}, q) |  \\
     & \leq \max_{q \in \mathcal{Q}} |L_w(\hat{w}_n, \frac{d^\pi_{P'}}{\mu}, q) | - \max_{q \in \mathcal{Q}} |L_{n, w}(\hat{w}_n, \frac{d^\pi_{P'}}{\mu}, q) | + \max_{q \in \mathcal{Q}} |L_{n,w}(\hat{w}_n, \frac{d^\pi_{P'}}{\mu}, q) | - \max_{q \in \mathcal{Q}} | L_w(\hat{w}, \frac{d^\pi_{P'}}{\mu}, q)|  
     + \\
     &\max_{q \in \mathcal{Q}} | L_w(\hat{w}, \frac{d^\pi_{P'}}{\mu}, q)| - \max_{q \in \mathcal{Q}}|L_w(\hat{w}, \hat{\beta}, q)|  + \max_{q \in \mathcal{Q}}|L_w(\hat{w}, \hat{\beta}, q)|  - \max_{q \in \mathcal{Q}}|L_w(\hat{w}, \frac{d^\pi_{P'}}{\mu}, q)| + \max_{q \in \mathcal{Q}}|L_w(\hat{w}, \frac{d^\pi_{P'}}{\mu}, q)| \\
          & \leq \max_{q \in \mathcal{Q}} |L_w(\hat{w}_n, \frac{d^\pi_{P'}}{\mu}, q) | - \max_{q \in \mathcal{Q}} |L_{n, w}(\hat{w}_n, \frac{d^\pi_{P'}}{\mu}, q) | + \max_{q \in \mathcal{Q}} |L_{n, w}(\hat{w}, \frac{d^\pi_{P'}}{\mu}, q) | - \max_{q \in \mathcal{Q}} | L_w(\hat{w}, \frac{d^\pi_{P}}{\mu}, q)| \\
     &
     + \max_{q \in \mathcal{Q}} | L_w(\hat{w}, \frac{d^\pi_{P'}}{\mu}, q)| - \max_{q \in \mathcal{Q}}|L_w(\hat{w}, \hat{\beta}, q)|  + \max_{q \in \mathcal{Q}}|L_w(\hat{w}, \hat{\beta}, q)| - \max_{q \in \mathcal{Q}}|L_w(\hat{w}, \frac{d^\pi_{P'}}{\mu}, q)| + \max_{q \in \mathcal{Q}}|L_w(\hat{w}, \frac{d^\pi_{P'}}{\mu}, q)|\nonumber  \\
     & \leq \underbrace{2\max_{q \in \mathcal{Q}, w \in \mathcal{W}} | |L_w(\hat{w}_n, \frac{d^\pi_{P'}}{\mu}, q) | - |L_n(\hat{w}_n, \frac{d^\pi_{P'}}{\mu}, q) | |}_{T1} + 2\underbrace{\max_{q \in \mathcal{Q}} | L_w(\hat{w}, \frac{d^\pi_{P'}}{\mu}, q) - L_w(\hat{w}, \hat{\beta}, q)|}_{T2} + \min_{w \in \mathcal{W}} \max_{q \in \mathcal{Q}} |L_w(w, \frac{d^\pi_{P_{tr}}}{\mu}, q)|
      \end{split}
 \end{align}
 Where, $\hat{w} = \arg \min_{w \in \mathcal{W}} \max_{q \in \mathcal{Q}} |L_w(w, \hat{\beta}, q)|$. Let's analyse each of the terms above one by one. Starting with T1 we get the following, 
 \begin{align}\label{eq:T1-w}
 \begin{split}
     T1 &= 2\max_{q \in \mathcal{Q}, w \in \mathcal{W}} | L_w(\hat{w}_n, \frac{d^\pi_{P'}}{\mu}, q) | - |L_{n, w}(\hat{w}_n, \frac{d^\pi_{P'}}{\mu}, q) | \\
     &\leq 2 \mathcal{R}_n(\mathcal{W}, \mathcal{Q}) + C_{\mathcal{W}} \cdot C_{\mathcal{Q}}\sqrt{\frac{\log (\frac{2}{\delta})}{2n}} \quad \text{w.p at-least $1-\frac{\delta}{2}$}
\end{split}
 \end{align}
 Where, the upper bound follows from ~\cite{DBLP:conf/colt/BartlettM01}. Note that $\mathcal{R}_n(\mathcal{W}, \mathcal{Q})$ is the Radamacher Complexity for the following function class 
 \begin{align}
 \begin{split}
\{ (s, a, s') \rightarrow w(s, a) \frac{d^\pi_{P_{tr}}(s, a)}{\mu(s, a)}(q(s, a) - \gamma q(s', \pi)): w \in \mathcal{W}, q \in \mathcal{Q}\} 
    \end{split}
\end{align}
 For the term T2 we can simplify the expression as follows, 
 \begin{align}\label{eq:T2-w}
 \begin{split}
     T2 &= 2\max_{q \in \mathcal{Q}} | L_w(\hat{w}, \frac{d^\pi_{P_{te}}}{\mu}, q) - L_w(\hat{w}, \hat{\beta}, q)| \\
     &= \max_{q \in \mathcal{Q}} |\mathbb{E}_{(s, a, s') \sim \mu}[ (\hat{\beta} - \frac{d^\pi_{P'}}{\mu}) \cdot \hat{w}(s, a) \cdot (q(s, a) - \gamma q(s', \pi))] | \\
     &= \max_{q \in \mathcal{Q}} |\mathbb{E}_{(s, a, s') \sim \mu}[ \varepsilon(s, a) \cdot \hat{w}(s, a) \cdot (q(s, a) - \gamma q(s', \pi))]|.
     \leq 2 C_{\mathcal{Q}} \cdot C_{\mathcal{W}} \|\varepsilon\|_\infty
     \end{split}
 \end{align}
 Here, we assume that $\varepsilon(s, a) = \hat{\beta} - \frac{d^\pi_{P'}}{\mu}$.Combining equations \ref{eq:T1-w}, \ref{eq:T2-w} along with equation \ref{eq:beta_error} we get the following upper-bound with at-least $1-\delta$
 \begin{align}
 \begin{split}
     &|\mathbb{E}_{(s, a) \sim d^\pi_{{P}}, r \sim R(s, a)} [\hat{w}_n\cdot r] - J_P(\pi)| \leq \max_{q \in \mathcal{Q}} |L_w(\hat{w}, \frac{d^\pi_{P_{tr}}}{\mu}, q)| + 4\gamma C_{\mathcal{W}} \cdot C_\mathcal{Q} \cdot  \|\varepsilon\|_{\infty} + 2 \mathcal{R}_n(\mathcal{W}, \mathcal{Q}) + C_{\mathcal{W}} \cdot C_{\mathcal{Q}}\sqrt{\frac{\log (\frac{2}{\delta})}{2n}}  \end{split}
 \end{align}
 Using Lemma \ref{le: estimation} we can bound $\|\varepsilon\|_{\infty}$ with probability $1 - \frac{\delta}{2}$ as follows, 
 \begin{align}
 \begin{split}
     &|\mathbb{E}_{(s, a) \sim d^\pi_{P}, r \sim R(s, a)} [\hat{w}_n \cdot r ] - J_P(\pi)|  \leq \min_{w \in \mathcal{W}}\max_{q \in \mathcal{Q}} |L_w(w, \frac{d^\pi_{P_{tr}}}{\mu}, q)| \\
     &+ 4 C_{\mathcal{W}} \cdot C_\mathcal{Q} \cdot \sqrt{2K \cdot \|\frac{d^\pi_{P_{tr}}}{\mu}\|_\infty \left(4\mathbb{E} \mathcal{R}_n(\mathcal{F}) + C_{\mathcal{F}}\sqrt{\frac{2 \log(\frac{2}{\delta})}{n}}\right)} + 4 \mathcal{R}_n(\mathcal{W}, \mathcal{Q}) + 2C_{\mathcal{W}} \cdot C_{\mathcal{Q}}\sqrt{\frac{\log (\frac{2}{\delta})}{2n}}  \\
     \end{split}
 \end{align}
 This completes the proof. \\
 \subsection{Proof of Theorem \ref{thm:finite-sample-2}}
Using equation \ref{eq:derivation-q}, we can bound the performance of the q estimator as follows, 
\begin{align}
\begin{split}
    &|(1-\gamma)\mathbb{E}_{d_0} [\hat{q}_n(s, \pi)] - J_P(\pi)| \leq \max_{w \in \mathcal{W}} | L_q(w, \frac{d^\pi_{P_{te}}}{\mu}, \hat{q}_n)| \\
    &\leq \max_{w \in \mathcal{W}} |L_q(w, \frac{d^\pi_{P_{te}}}{\mu}, \hat{q}_n)| - \max_{w \in \mathcal{W}} |L_{n, q}(w, \frac{d^\pi_{P_{te}}}{\mu}, \hat{q}_n)| + \max_{w \in \mathcal{W}} |L_{n, q}(w, \frac{d^\pi_{P_{te}}}{\mu}, \hat{q}_n)| - \max_{w \in \mathcal{W}} |L_q(w, \frac{d^\pi_{P_{tr}}}{\mu}, \hat{q})| \\
    & + \max_{w \in \mathcal{W}} |L_q(w, \frac{d^\pi_{P_{tr}}}{\mu}, \hat{q})| - \max_{w \in \mathcal{W}} |L_q(w, \hat{\beta},\hat{q})| + \max_{w \in \mathcal{W}} |L_q(w, \hat{\beta}, \hat{q})| - \max_{w \in \mathcal{W}} |L_q(w, \frac{d^\pi_{P_{tr}}}{\mu}, \hat{q})| + \max_{w \in \mathcal{W}} |L_q(w, \frac{d^\pi_{P_{tr}}}{\mu}, \hat{q})|\\
    &\leq  \max_{w \in \mathcal{W}} |L_q(w, \frac{d^\pi_{P_{te}}}{\mu}, \hat{q}_n)| - \max_{w \in \mathcal{W}} |L_{n, q}(w, \frac{d^\pi_{P_{te}}}{\mu}, \hat{q}_n)| + \max_{w \in \mathcal{W}}|L_{n, q}(w, \frac{d^\pi_P}{\mu}, \hat{q}_n)| - \max_{w \in \mathcal{W}} |L_q(w, \frac{d^\pi_{P_{tr}}}{\mu}, \hat{q}_n)| \\
    &+ \max_{w \in \mathcal{W}} |L_q(w, \frac{d^\pi_{P_{tr}}}{\mu}, \hat{q})| - \max_{w \in \mathcal{W}} |L_q(w, \hat{\beta}, \hat{q})| + \max_{w \in \mathcal{W}} |L_q(w, \hat{\beta}, \hat{q})| - \max_{w \in \mathcal{W}} |L_q(w, \frac{d^\pi_{P_{tr}}}{\mu}, \hat{q})| + \max_{w \in \mathcal{W}} |L_q(w, \frac{d^\pi_{P_{tr}}}{\mu}, \hat{q})|\\
    &\leq 2 \underbrace{\max_{q \in \mathcal{Q}, w \in \mathcal{W}} |L_q(w, \frac{d^\pi_{P_{te}}}{\mu}, q) -  L_{n, q}(w, \frac{d^\pi_{P_{te}}}{\mu}, q)|}_{T1} + 2\underbrace{\max_{w \in \mathcal{W}}|L_q(w, \frac{d^\pi_{P_{tr}}}{\mu}, \hat{q}) - L_q(w, \hat{\beta}, \hat{q})|}_{T2} + \min_{q \in \mathcal{Q}} \max_{w \in \mathcal{W}} L_q(w, \frac{d^\pi_{P_{tr}}}{\mu}, q)
    \end{split}
\end{align}
Where, $\hat{q} = \arg \min_{q \in \mathcal{Q}} \max_{q \in \mathcal{W}} |L_q(w, \hat{\beta}, q)|$. Lets analyse each of these terms $T1, T2$ separately. For T1 we get the following, 
 \begin{align}\label{eq:T1_q}
 \begin{split}
     T1 &= 2\max_{q \in \mathcal{Q}, w \in \mathcal{W}} |L_q(w, \frac{d^\pi_{P_{te}}}{\mu}, q) -  L_{n, q}(w, \frac{d^\pi_{P_{te}}}{\mu}, q)|\\
     &\leq 2 \mathcal{R}_n(\mathcal{W}, \mathcal{Q}) + C_{\mathcal{W}} \cdot \frac{R_{\max}}{1-\gamma}\sqrt{\frac{\log (\frac{2}{\delta})}{2n}} \quad \text{w.p at-least $1-\frac{\delta}{2}$}
\end{split}
\end{align}
 Where, the upper bound follows from ~\cite{DBLP:conf/colt/BartlettM01}. Note that $\mathcal{R}_n(\mathcal{W}, \mathcal{Q})$ is the Radamacher Complexity for the following function class 
 \begin{align}
 \begin{split}
\{ (s, a, s') \rightarrow w(s, a) \frac{d^\pi_{P_{tr}}(s, a)}{\mu(s, a)}(q(s, a) - \gamma q(s', \pi)): w \in \mathcal{W}, q \in \mathcal{Q}\} 
    \end{split}
\end{align}
 For the term T2 we can simplify the expression as follows, 
 \begin{align}\label{eq:T2_q}
 \begin{split}
     T2 &= 2\max_{w \in \mathcal{W}} | L_q(w, \frac{d^\pi_{P_{tr}}}{\mu}, \hat{q}) - L_q(w, \hat{\beta}, \hat{q})| \\
     &= \max_{w \in \mathcal{W}} |\mathbb{E}_{(s, a, s') \sim \mu}[ (\hat{\beta} - \frac{d^\pi_{P'}}{\mu}) \cdot w(s, a) \cdot (\hat{q}(s, a) - \gamma \hat{q}(s', \pi)] | \\
     &\leq 2C_{\mathcal{W}} \frac{R_{max}}{1 - \gamma} \|\varepsilon\|_{\infty}
     \end{split}
 \end{align}
Combining equation \ref{eq:T1_q} and \ref{eq:T2_q} along with equation \ref{eq:beta_error} we can bound the error in evaluation as follows, 
\begin{align}
\begin{split}
    &|(1-\gamma)\mathbb{E}_{d_0} [\hat{q}_n(s, \pi)] - J_P(\pi)| \leq \min_{q \in \mathcal{Q}} \max_{w \in \mathcal{W}} L_q(w, \frac{d^\pi}{\mu}, q) +  2 \mathcal{R}_n(\mathcal{W}, \mathcal{Q}) + 4C_{\mathcal{W}} \cdot \frac{R_{\max}}{1-\gamma}\sqrt{\frac{\log  (\frac{2}{\delta})}{2n}}   \\
    &+  2C_{\mathcal{W}} \frac{R_{max}}{1 - \gamma} \sqrt{2K \cdot \|\frac{d^\pi_{P_{tr}}}{\mu} \|_\infty \left(4\mathbb{E} \mathcal{R}_n(\mathcal{F}) + C_{\mathcal{F}}\sqrt{\frac{2 \log(\frac{2}{\delta})}{n}}\right)}\\
    & \quad \text{w.p at-least $1-\delta$}
    \end{split}
\end{align}
This completes the proof
\subsection{Additional Literature Review}
Reinforcement learning and its application to robotics has garnered considerable interest from the robotics community ~\cite{sikchi2023imitation, DBLP:journals/corr/abs-2001-01866}. A key aspect of reinforcement learning applied to robotics problems is our inability to train a robot in the real-world for a long period of time. To that end, four different themes of reinforcement learning approaches have developed namely, Sim2Real, Transfer Learning, Imitation Learning, and Offline Reinforcement Learning. 
\begin{enumerate}
    \item \textbf{Sim2Real:} algorithms are primarily concerned with learning robust policy in simulation by training the algorithms over a variety of simulation configurations ~\cite{DBLP:journals/tase/HoferBHGMGAFGLL21, DBLP:conf/iros/TobinFRSZA17, DBLP:journals/corr/abs-1903-11774}. Sim2Real algorithms, although successful, still require a thorough real-world deployment in order to gauge the policy’s performance
    \item \textbf{Transfer learning} algorithms on the other hand involve an agent’s interaction with the real world, but in a limited manner. Transfer learning algorithms typically involve back and forth between the real world and the simulator, with the caveat that interaction with the real-world is kept limited ~\cite{DBLP:journals/corr/WulfmeierPA17, DBLP:conf/iclr/EysenbachCALS21, DBLP:journals/corr/abs-2009-07888}. Since, transfer learning algorithms require multiple deployment in the real world to learn. There arises safety critical issues when attempting to learn optimal policies. 
    \item  \textbf{Imitation learning} algorithms learn an optimal policy by trying to mimic offline expert demonstrations ~\cite{DBLP:journals/jmlr/RossGB11, DBLP:conf/nips/HoE16, DBLP:journals/corr/abs-1901-09387}. Many successful imitation learning algorithms minimize some form of density matching between the expert demonstrations and on-policy data to learn optimal policy. A key problem with imitation learning is the fact that it requires constant interaction with the real-world environment in order to learn an optimal policy. 
    \item \textbf{Offline reinforcement learning} is a relatively new area. Here the idea is to learn an optimal policy using offline data without any interaction with the environment ~\cite{kumar2020conservative, DBLP:journals/corr/abs-2106-10783, DBLP:conf/nips/YuKRRLF21, DBLP:conf/aaai/YuYSNE23}. Offline reinforcement learning has recently demonstrated performance at-par with classical reinforcement learning in a few tasks. However, offline reinforcement learning algorithms tend to overfit on the offline data. Thus, even offline learning modules too require an actual deployment in-order to assess performance. 
\end{enumerate}
\subsection{Additional Experimental Details and Additional Results}\label{sec:experimental_details}
\begin{figure}[!htb]
\begin{subfigure}{\textwidth}
    \centering
    \includegraphics[scale=0.6]{outline/results/cartpole/legend.png}
\end{subfigure}\hfill
\begin{subfigure}{\textwidth}
    \centering
    \includegraphics[width=\textwidth]{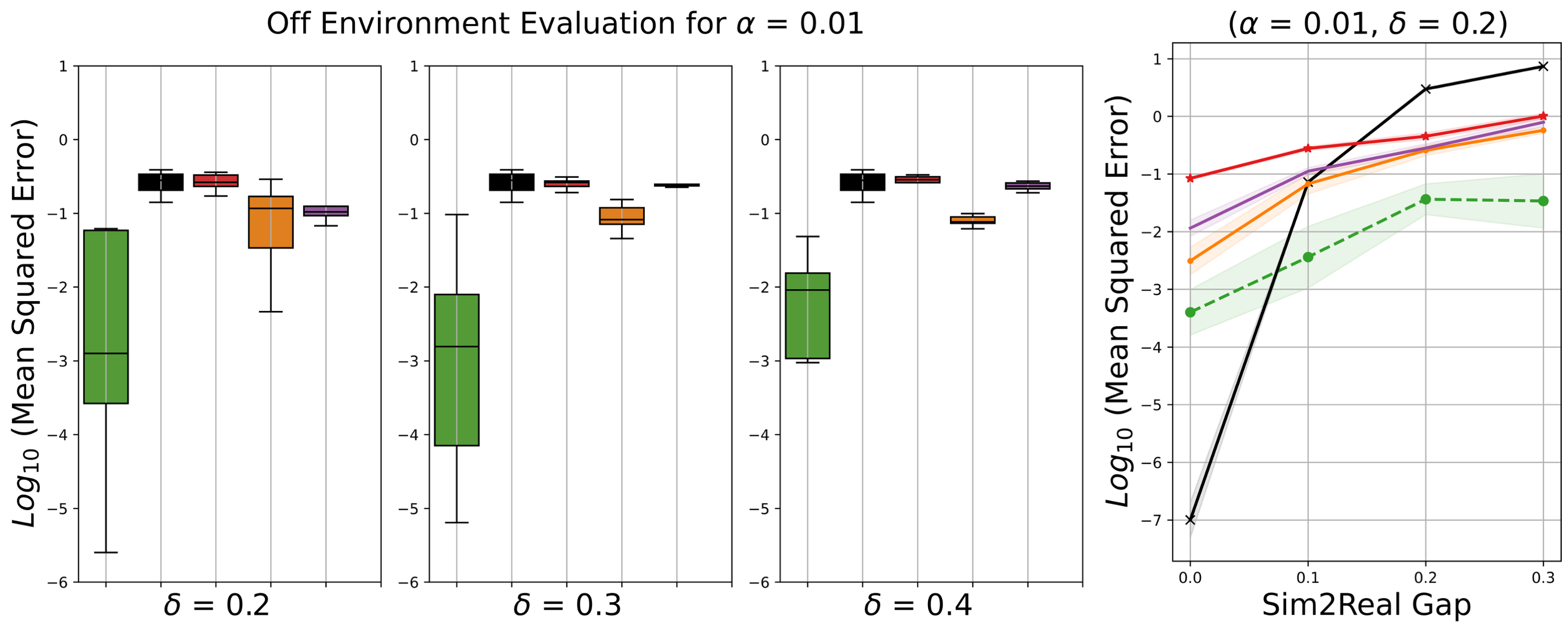}
    \caption{$\alpha=0.01$}
\end{subfigure}\hfill
\begin{subfigure}{\textwidth}
    \centering
    \includegraphics[width=\textwidth]{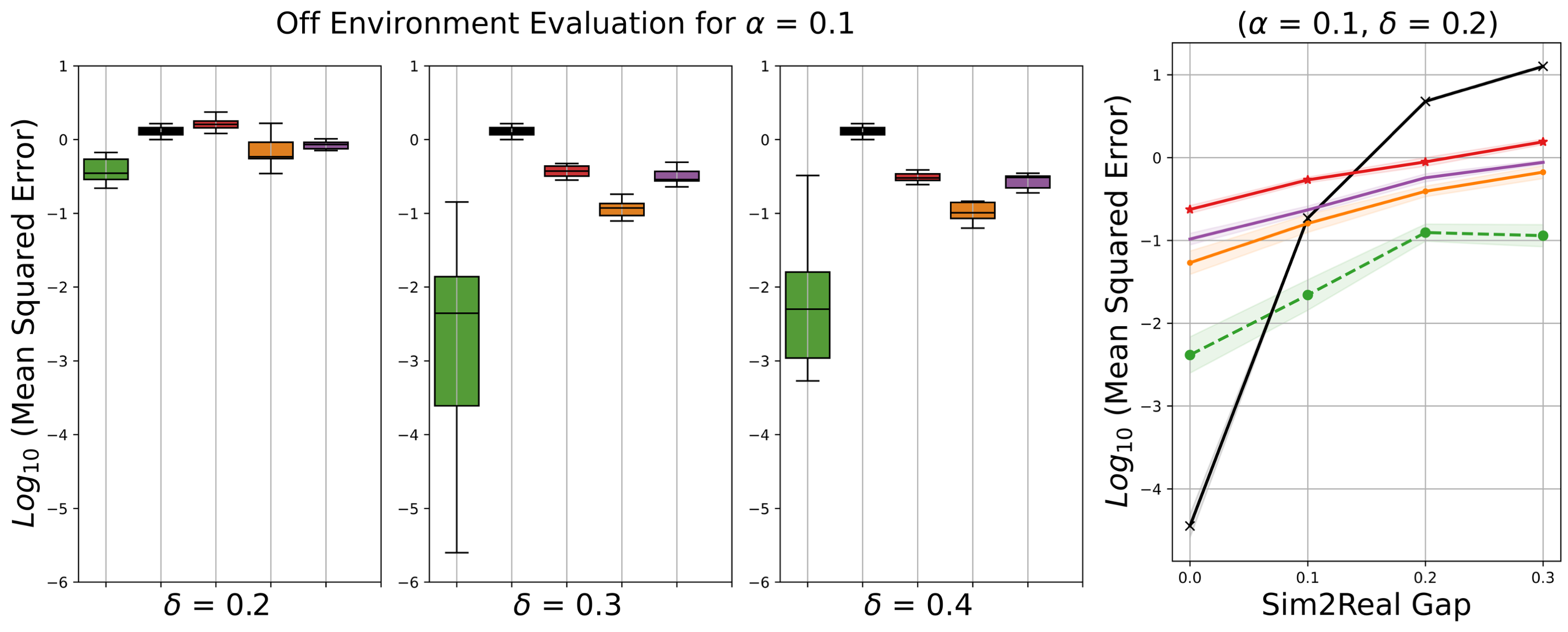}
    \caption{$\alpha=0.1$}
\end{subfigure}\hfill
\begin{subfigure}{\textwidth}
    \centering
    \includegraphics[width=0.7\textwidth]{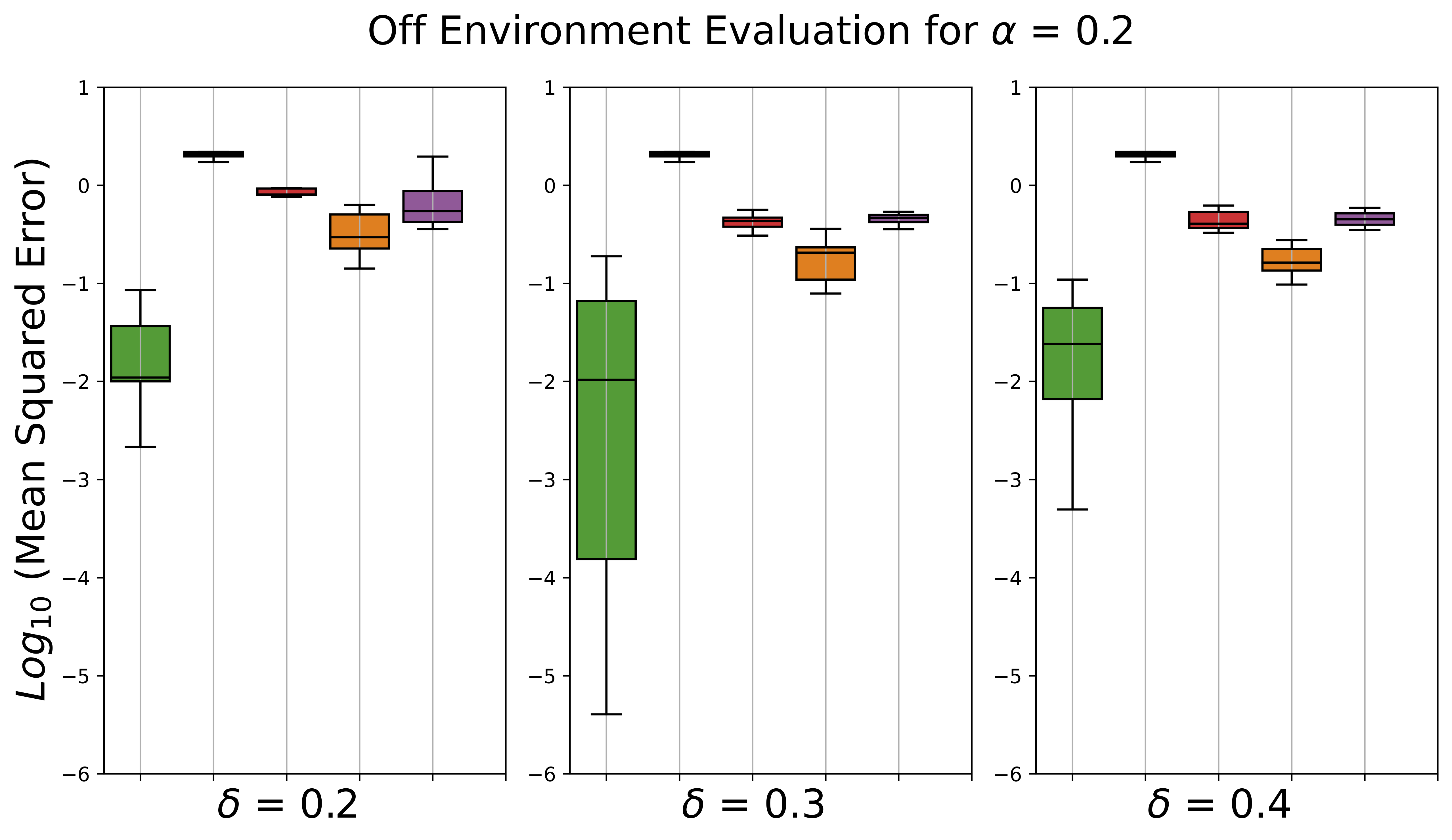}
    \caption{$\alpha=0.2$}
\end{subfigure}\hfill
    \caption{Each of the above figure demonstrates the effect of evaluation over varying behavior policies $\delta =\{0.2, 0.3, 0.4\}$ on a fixed target policy using $\beta$-DICE for the taxi environment. For these set of experiments the training environment is the default transition parameters, while the test environment has $\tau = 0.1$. In (a), (b), (c) the target policies that we use are $\alpha = \{0.01$, $0.1, 0.2$\}. Additionally for (a), (b) (RHS) we also show the effect of varying sim2real gap on target policy evaluation using $\beta$-DICE (while keeping  $\delta, \alpha$ fixed).}
    \label{fig:additional_taxi}
\end{figure}

\begin{figure}
\begin{subfigure}{\textwidth}
    \centering
    \includegraphics[scale=0.6]{outline/results/cartpole/legend.png}
\end{subfigure}\hfill
\begin{subfigure}{\textwidth}
    \centering
    \includegraphics[width=\textwidth]{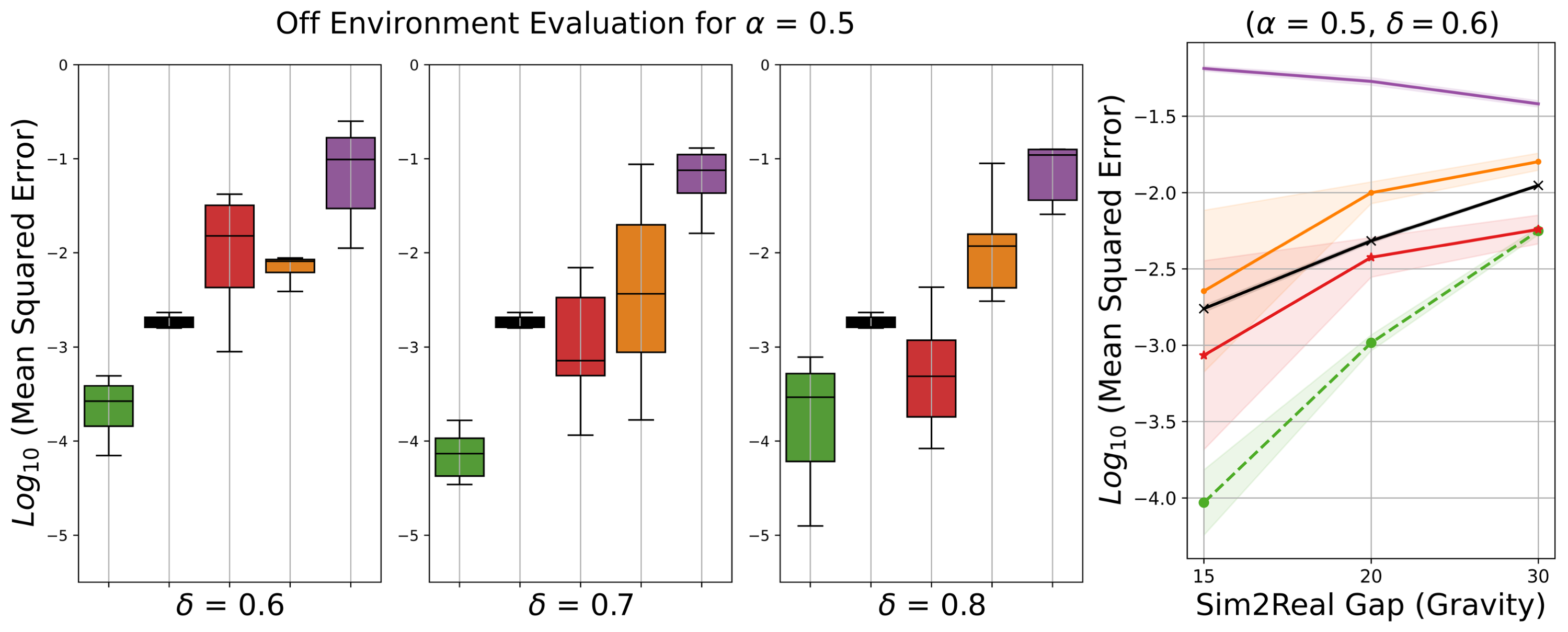}
    \caption{$\alpha=0.5$}
\end{subfigure}\hfill
\begin{subfigure}{\textwidth}
    \includegraphics[width=0.7\textwidth]{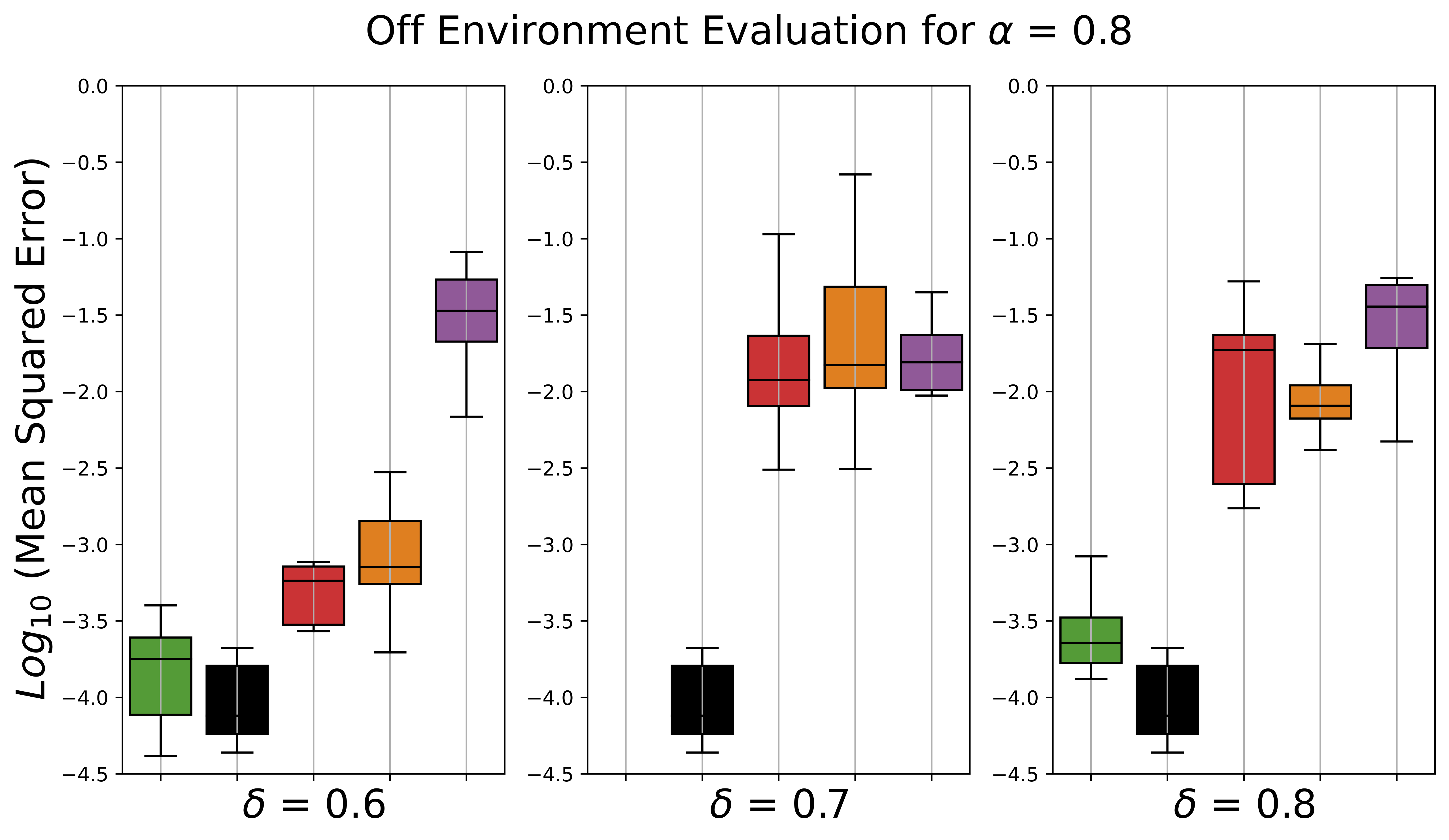}
    \caption{$\alpha=0.8$}
\end{subfigure}
    \caption{Each of the above figure demonstrates the effect of evaluation over varying behavior policies $\delta =\{0.6, 0.7, 0.8\}$ on a fixed target policy using $\beta$-DICE for the cartpole environment. For these set of experiments the training environment has gravity = $10 m/s^2$, while the test environment has gravity = $15.0 m/s^2$. In (a), (b) the target policies that we use are $\alpha = \{0.5, 0.8$\}. Additionally for (a), (RHS) we also show the effect of varying sim2real gap on target policy evaluation using $\beta$-DICE (while keeping  $\delta, \alpha$ fixed).}
    \label{fig:additional_cartpole}
\end{figure}

\begin{figure}
\begin{subfigure}{\textwidth}
    \centering
    \includegraphics[scale=0.6]{outline/results/cartpole/legend.png}
\end{subfigure}\hfill
\begin{subfigure}{\textwidth}
    \centering
    \includegraphics[width=\textwidth]{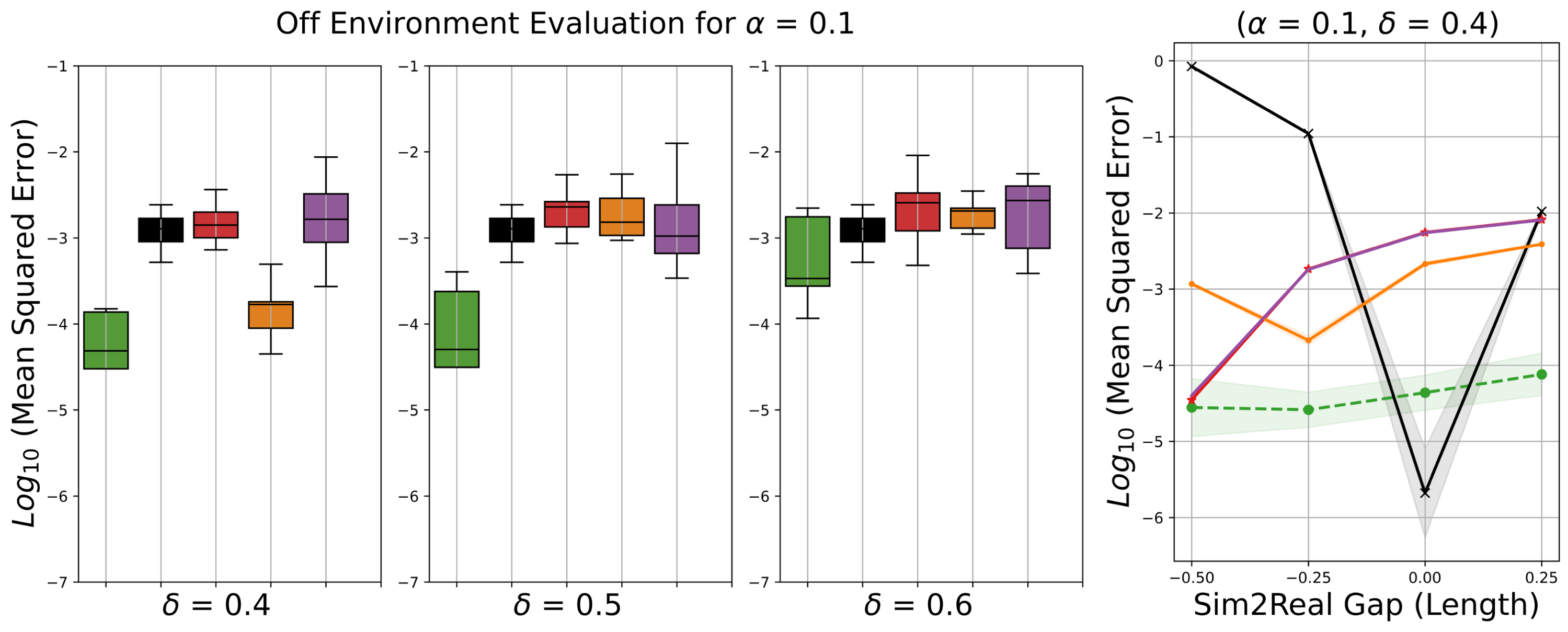}
    \caption{$\alpha=0.1$}
    \label{fig:additional_reacher_a}
\end{subfigure}\hfill
\begin{subfigure}{\textwidth}
    \centering
    \includegraphics[width=0.7\textwidth]{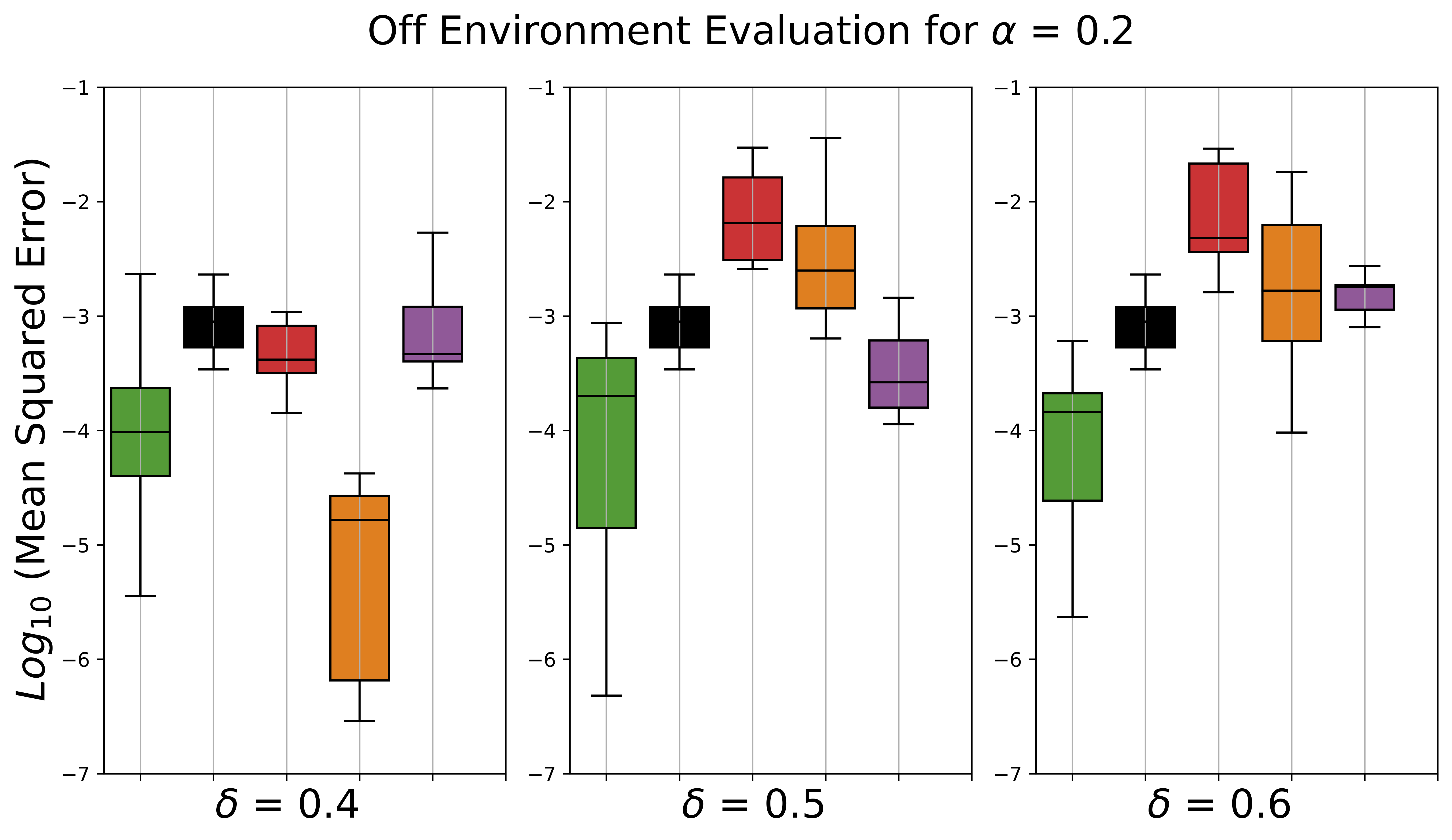}
    \caption{$\alpha=0.2$}
\end{subfigure}
    \caption{Each of the above figure demonstrates the effect of evaluation over varying behavior policies $\delta =\{0.4, 0.5, 0.6\}$ on a fixed target policy using $\beta$-DICE for the reacher environment. For these set of experiments the training environment has length = $0.1 m$, while the test environment has length = $0.075 m$. In (a), (b) the target policies that we use are $\alpha = \{0.1, 0.2$\}. Additionally for (a), (RHS) we also show the effect of varying sim2real gap on target policy evaluation using $\beta$-DICE (while keeping  $\delta, \alpha$ fixed).}
    \label{fig:additional_reacher}
\end{figure}

\begin{figure}
\begin{subfigure}{\textwidth}
    \centering
    \includegraphics[scale=0.6]{outline/results/cartpole/legend.png}
\end{subfigure}\hfill
\begin{subfigure}{\textwidth}
    \centering
    \includegraphics[width=0.7\textwidth]{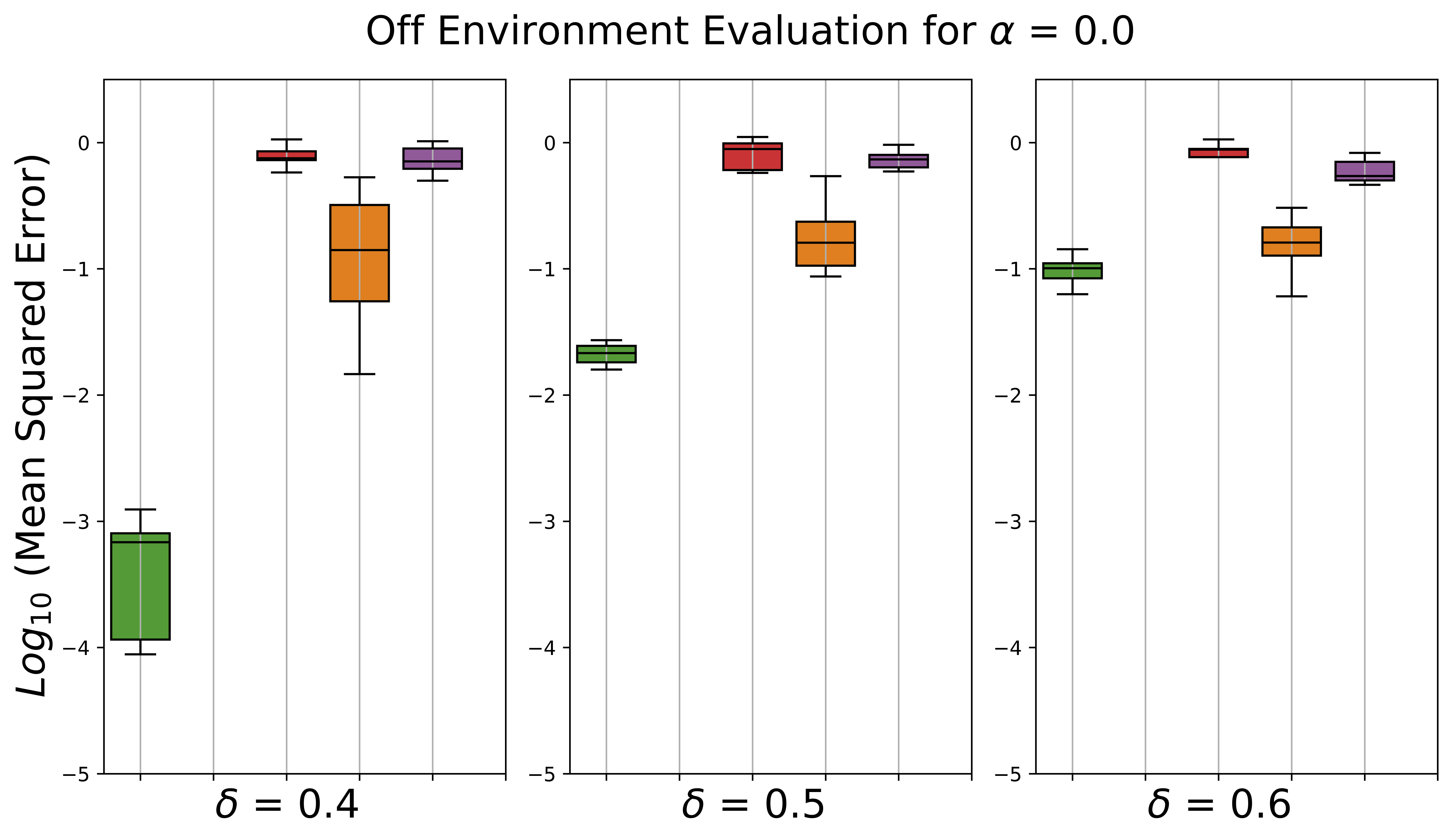}
    \caption{$\alpha=0.0$}
\end{subfigure}\hfill
\begin{subfigure}{\textwidth}
    \centering
    \includegraphics[width=0.7\textwidth]{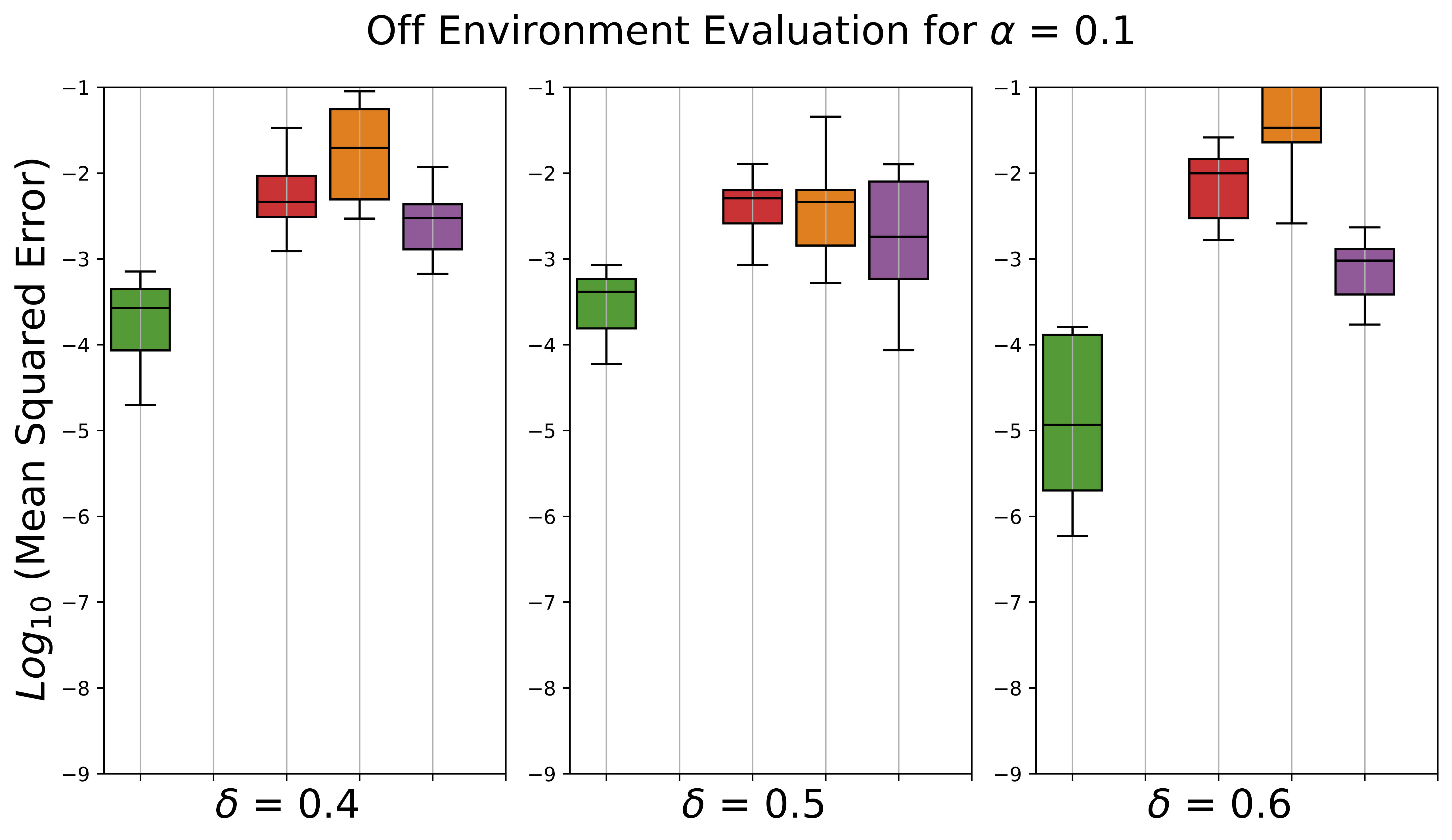}
    \caption{$\alpha=0.1$}
\end{subfigure}\hfill
\begin{subfigure}{\textwidth}
    \centering
    \includegraphics[width=0.7\textwidth]{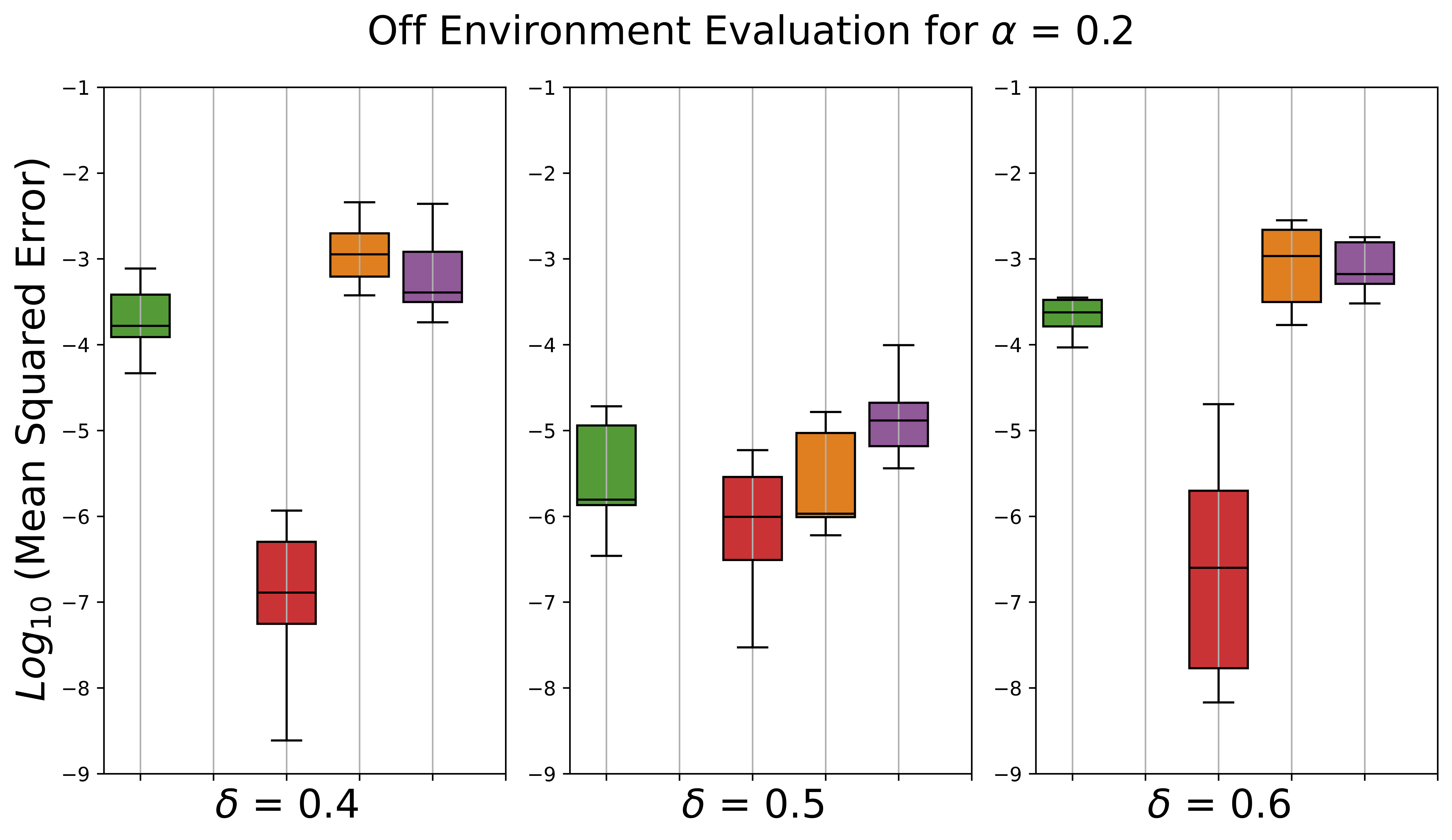}
    \caption{$\alpha=0.2$}
\end{subfigure}\hfill

    \caption{Each of the above figure demonstrates the effect of evaluation over varying behavior policies $\delta =\{0.4, 0.5, 0.6\}$ on a fixed target policy using $\beta$-DICE for the half cheetah environment. For these set of experiments the training environment has length = $0.9 Nm$, while the test environment has length = $1.3 Nm$. In (a), (b), (c) the target policies that we use are $\alpha = \{0.0, 0.1, 0.2$\}}
    \label{fig:halfcheetah}
\end{figure}

\begin{table}[t]
\captionsetup{singlelinecheck = false, justification=justified}
\captionof{table}{Configuration for various Sim2Sim environment pair over which our algorithm was deployed. Here, we define $\varepsilon$ for taxi environment, environment gravity ($g$) for cartpole environment, both the link lengths ($l$) for reacher environment and maximum torque power ($p$) for half-cheetah environment. Note that $\varepsilon$ is the amount of random noise injected into the default taxi environment.}
\normalsize
\label{tab:sim2sim_parameters}
\begin{center} 
\begin{tabular}{lccc}
\toprule 
Environments   &Parameter changed & Sim Environment & "Real" Environment\\
\midrule
{\tt Taxi}~\cite{brockman2016openai}  & $\varepsilon$ & $\varepsilon=0.0$ & $\varepsilon=\{0.1, 0.2, 0.3\}$\\
{\tt Cartpole}~\cite{brockman2016openai}  &$g (m/s^2)$& $g=-9.81$ & $g=\{-15, -20, -30\} $\\
{\tt Reacher}~\cite{DBLP:journals/corr/SchulmanWDRK17}  &$l(m)$& $l = 0.1$ & $l=\{5, 7.5, 10, 12.5\} 10e^{-2}$\\
{\tt HalfCheetah}~\cite{PackerGao:1810.12282}  &$p(N.m)$& $p = 0.9$ & $p=\{1.3\}$\\
\bottomrule
\end{tabular}
\end{center}
\vspace{-15pt}
\end{table}
\textbf{Experiment Setup} 
We conduct experiments on both Sim2Sim and Sim2Real environments. For Sim2Sim experiments we demonstrate our results over a range of different types of environments like Tabular (Taxi), Discrete-control (cartpole) and continuous control (Reacher and Halfcheetah). For the Sim2Sim experiments over a diverse set of simulation and the real world environments like gravity, arm-length, friction and maximum torque. For all the experiments we mention here, we will first generate an offline data which was collected using known behavior policy $\mu$. For the sake of these experiments, behavior policy are parameterized by a factor $\delta$ which basically dictates the amount of noise added to a pre-trained model. We similarly parameterise the target policy target policy by $\alpha$. We experiment over different pairs of training and test environments. We typically keep the simulation environment fixed and vary the test environment.  We call the key parametric difference between the training and the test environment as the Sim2Real gap. Detailed information for each set of experiments is provided below. 
\\
\textbf{Learning $\beta$}: We parameterize $\beta$ as two-layered neural network with ReLU activation layers for intermediate layers. We experimented with two different kinds of final activation layer, squared and tanh. We observed that tanh layer scaled to go from 0 to 10 worked best for these set of experiments. \\
\textbf{Learning $w$}: For most of our experiments on $\beta$-DICE we use the framework of GradientDICE. GradientDICE algorithms are typically two layered neural networks which use orthogonal initialisation. Inner activation is ReLU and the final activation layer is linear. \\
\textbf{Baselines} We compare with the following baselines:
\begin{itemize}
\item Simulator: This is the baseline of trusting the simulator's evaluation and not using data from the test environment.
\item Model-free MIS: We include DualDICE, GradientDICE, GenDICE ~\cite{DBLP:journals/corr/abs-1906-04733, zhang2019gendice, zhang2020gradientdice} as state-of-the-art baselines for model-free MIS, which only uses data from the test environment and does not use simulator information.
\item Residual dynamics: We fit a model for OPE from test-environment data with the simulator as the ``base'' prediction and only learn a correction term. 
\item DR-DICE \cite{tang2019doubly}: the previous baselines ignore some of the available information (e.g., model-free MIS does not use simulator information) or use them in a na\"ive manner. Therefore, we additionally include a doubly-robust (DR) MIS estimator \cite{tang2019doubly} that can organically blend the model information with the test-environment data. 
\end{itemize}
\textbf{Taxi Environment:} Taxi environment has 500 states and 6 discrete actions. For these set of experiments the simulator environment involves deterministic transition between two states. For the real world environment, we experiment with environments where the transition is deterministic with probability $(1-\tau)$ and random with probability $\tau$. With $\tau$ being the Sim2Real gap. To collect data, we use a behavior policy that chooses optimal action (which was learnt using Q-learning) with probability $1-\delta$ and a random action with probability $\delta$. Target policies are similarly parameterised but with $\alpha$. In figure \ref{fig:additional_taxi} we demonstrate the performance of $\beta$-DICE for $\alpha = 0.1$. In these set of experiments, we evaluate performance of $\beta$-DICE over 3 different types of behavior policies $\delta = \{0.2, 0.3, .0.4\}$ and three different sets of target policies $\alpha = \{0.01, 0.1, 0.2\}$. For two sets of behavior and target policy, we also show the effect of sim gap on evaluation error. We observe that evaluation error increases with increasing sim2sim gap. 
For these set of experiments we used a discounting factor $\gamma = 0.9$ and limited our offline trajectory collection to 150 timesteps. Learning rate for $\beta$ is 1e-4, the learning rate for $w$ is 1e-4. We observe that $\beta$-DICE is able to outperform the state-of-the-art MIS baseline comfortably. \\
\textbf{Cartpole Environment:} For discrete control problems, we choose the Cartpole environment ~\cite{1606.01540}. For the simulator we choose cartpole environment with gravity equals to $10 m/s^2$. For the test environment, we choose gravity to be $(\tau) m/s^2$. With $\tau$ being the Sim2Sim gap. Our behavior policy is chosen to be a mixture of optimal policy (which was trained using Cross Entropy method) $\pi_*$ and a uniformly random policy $U$ such that $\mu = (1-\delta)\pi_* + (\delta)U$. Our target policy is similarly parameterised by $\alpha$. We demonstrate results over different sets of behavior policies $\delta = \{0.4, 0.5, 0.6\}$ and evaluate performance over a set of $\alpha=\{0.2, 0.5, 0.8\}$ and simreal gap $\tau =\{5, 10, 20\} m/s^2$. In figures \ref{fig:cartpole} and \ref{fig:additional_cartpole} (with additional baselines), we demonstrate our experiments over different sets of behavior policies and target policies and observe that our method is more than capable of improving upon state-of-the-art baseline with information from simulation. Our discounting factor $\gamma = 0.99$ and timesteps is limited to 200. Learning rate for $\beta$ is 1e-4 and learning rate for $w$ is 1e-2. \\
\textbf{Reacher Environment:} For continuous control, we experiment with RoboschoolReacher environment. For these set of environments, we choose training environment as the one where the length of both links are 0.1 m. The test environment is chosen to be one, where the length of both the links is $(0.1 + \tau) m$. We choose behavior policy as the addition of an optimal policy plus a zero mean normal policy whose standard deviation is $\delta$. For our experiments, $\delta = \{0.4, 0.5, 0.6 \}$, $\alpha=\{0.0, 0.1, 0.2\}$ and $\tau = \{-0.5, -0.25, 0.0, 0.25\} m$. In figures \ref{fig:reacher} and \ref{fig:additional_reacher}, we demonstrate our experiments over different sets of behavior policies and target policies and observe that our method is more than capable of improving upon state-of-the-art baseline with information from simulation. In figure \ref{fig:reacher} and \ref{fig:additional_reacher_a}, we also demonstrate the effect of $\beta$-DICE with sim2sim gap over two sets of $(\delta, \alpha)$. Our discounting factor $\gamma = 0.99$ and timesteps is limited to 150. Learning rate for $\beta$ is 1e-4 and learning rate for $w$ is 3e-3.\\
\textbf{HalfCheetah Environment:} For continuous control, we experiment with RoboschoolCheetah environment. For these set of environments, we choose training environment as the one where the maximum torque to the joints is $0.9$. The test environment is chosen to be one, where the length of both the links is $0.9 + \tau$ N.m We choose behavior policy as the addition of an optimal policy with zero mean normal policy whose standard deviation is $\delta$. For behavior policy the delta is taken to be $\delta = \{0.4, 0.5, 0.6 \}$ and the target policy is taken to be $\alpha=\{0.0, 0.1, 0.2\}$. Due to limited computation, we experimented only with $\tau = 0.4 Nm$. In figures \ref{fig:halfcheetah}, we demonstrate our experiments over different sets of behavior policies and target policies and observe that our method is more than capable of improving upon state-of-the-art baseline with information from simulation. Our discounting factor $\gamma = 0.99$ and timesteps is limited to 150. Learning rate for $\beta, w$ is 1e-4.\\


\end{document}